\documentclass[preprint,12pt]{elsarticle}

\journal{Enterprise Information Systems}

\usepackage{epstopdf}
\usepackage{subfigure}
\usepackage{hyperref}
\usepackage{float}

\usepackage{amssymb}

\usepackage{multirow}
\usepackage[table,xcdraw]{xcolor}
\usepackage{tabularx}
\usepackage{lipsum} 
\usepackage[utf8]{inputenc}
\usepackage{geometry}
\usepackage{hyperref}
\usepackage{booktabs} 
\usepackage{makecell} 
\usepackage{hyperref}
\usepackage[table]{xcolor} 
\usepackage{array}

\usepackage{threeparttable}
\usepackage{natbib}
\setcitestyle{square}
\usepackage{caption}
\usepackage{boldline}
\usepackage{makecell}
\usepackage{marvosym}
\usepackage{rotating}
\usepackage{multicol}
\usepackage{multirow}
\usepackage{longtable}

\begin{document}


\begin{frontmatter}



\title{NLP4PBM: A Systematic Review on Process Extraction using Natural Language Processing with Rule-based, Machine and Deep Learning Methods}

\author[inst1]{William Van Woensel\corref{c1}}

\affiliation[inst1]{organization={Telfer School of Management, University of Ottawa},
            addressline={55 Laurier E.}, 
            city={Ottawa},
            postcode={K1N 6N5}, 
            state={Ontario},
            country={Canada}}

\cortext[c1]{Corresponding author}
\ead{wvanwoen@uottawa.ca}

\author[inst1]{Soroor Motie}


\begin{abstract}
This literature review studies the field of automated process extraction, i.e., transforming textual descriptions into structured processes using Natural Language Processing (NLP).
We found that Machine Learning (ML) / Deep Learning (DL) methods are being increasingly used for the NLP component. In some cases, they were chosen for their suitability towards process extraction, and results show that they can outperform classic rule-based methods. We also found a paucity of gold-standard, scalable annotated datasets, which currently hinders objective evaluations as well as the training or fine-tuning of ML / DL methods. 
Finally, we discuss preliminary work on the application of LLMs for automated process extraction, as well as promising developments in this field.
\end{abstract}


\end{frontmatter}

\section{Introduction}
Business processes 
are structured as series of tasks that collectively accomplish a business objective \cite{weske2007business}. 
As such, they can provide an objective framework for an organization's day-to-day operations \cite{weske2007business}. 
Business Process Management (BPM) is devoted to the analysis, design, implementation and management of business processes \cite{weske2007business}. BPM commonly relies on process models, i.e., structured representations of the processes in question~\cite{bellan2020qualitative}. 
Process models may represent control flow (e.g., sequence, concurrency) or decisional (e.g., criteria and their requirements) aspects of processes. 
Moreover, in the former category, process models can be imperative, i.e., exhaustively prescribing the occurrence and ordering of activities;
or declarative, i.e., declaring high-level constraints on the occurrence and ordering of specific activities\footnote{E.g., a \textit{precedence} constraint on (A, B) states that B only occurs if preceded by A.}~\cite{wvda_declare}.

Currently, most organizations rely solely on natural text to describe business processes and their requirements~\cite{van2018challenges}. Vast amounts of textual data relevant to business processes is constantly being generated, such as reports, emails, and ad-hoc documentation. 
To implement BPM in such organizations, textual sources could be used as a source for structured process models.
The extraction of process models
from natural language text, albeit manually or automatically, is referred to as \textit{process extraction}~\cite{bellan2020qualitative}. 

Nevertheless, according to a 2019 report from Deloitte~\cite{Davenport2019}, only a few organizations
(18\%) analyze unstructured data such as natural text to gain business insights, including as the processes described therein, since it is more challenging to interpret.
Indeed, manual process extraction in particular is known to be time-consuming and error-prone~\cite{maqbool2019comprehensive,bordignon2018natural,bellan2020qualitative}; Herbst reported that more than half of the time in BPM is often spent on manual process extraction~\cite{herbst_inductive_1999}.
Hence, automated support for process extraction, if done accurately and reliably, can improve the efficiency of process extraction. 
To that end, a number of studies have used Natural Language Processing (NLP) methods to implement process extraction: we refer to this as \textit{automated process extraction}, or simply \textit{process extraction} if the automated aspect is clear from the context.
By providing structured process models, automated process extraction thus has the potential to improve the effectiveness of BPM in organizations currently relying on textual data. 
The availability of extracted process models, i.e., their management, storage and retrieval, will further have implications for information management within enterprises.

Traditionally, process extraction methods have involved rule-based and/or Machine Learning(ML)-based methods to implement two steps:
(1)~\textit{NLP}, which categorises the fundamental building blocks of the text; such as labeling nouns, verbs, and adjectives, recognizing unique entities (e.g., people, events), and tagging their semantic concepts (e.g., actor, activity); and
(2)~\textit{process generation}, which maps the NLP output to a process model that captures the control-flow or decisional elements from the text. 
With the emergence of Deep Learning (DL) in NLP, the last 5 years have seen a paradigm shift: authors are increasingly using DL models such as Transformers (e.g., BERT \cite{goossens2023extracting}) and Long Short-Term Memory (LSTM) \cite{han2020bps} in process extraction.
Even more recently, the advent of Large Language Models (LLM) has led to a surge of interest in LLM-powered NLP for a number of fields. 
We expect LLM to also impact the process extraction field, given their capabilities to generate code and models \cite{kampik2023large}; we have already seen preliminary work on this topic~\cite{bellan_gpt3_2022,grohs2023large,forell_modeling_meets_llm,kourani_process_modelling_with_llm}.
We discuss this initial work, and the promise of recent innovations, in the Discussion.

With these promising natural language applications on the horizon, 
we believe it is opportune to present a systematic review of process extraction from the ``pre-LLM'' era. 
For researchers who want to contribute to this field, 
we aim to acquaint them with state-of-the-art methods and pipelines, applied evaluation methods and datasets, and preliminary work on LLM in process extraction.
For practitioners who want to apply process extraction, we present an overview of relevant methods, including publicly available NLP tools.
We discuss both traditional approaches to process extraction, and emphasise recent work that relies on ML/DL.
Hence, this paper will cover the following research questions:
\begin{itemize}
\item RQ1: Which NLP and process generation methods, including publicly available tools, have been studied and used for extracting process models?
\item RQ2: To what extent have ML/DL methods been studied in process extraction?
\item RQ3: Which evaluation methods and datasets have been utilised to evaluate process extraction?
\end{itemize}

The rest of the paper is structured as follows. Section~\ref{sec:rel_work} discusses other review papers on automated process extraction. Section~\ref{sec:methods} outlines our systematic review approach, including source databases, keywords, exclusion and inclusion criteria, and classification criteria. Sections \ref{sec:results}-\ref{sec:results_eval} present our data synthesis results, including answers to the research questions. Section~\ref{sec:discussion} summarises important observations from our study.
Section~\ref{sec:threats_limits} discusses the limitations and threats to the validity of the study, and Section~\ref{sec:conclusion} concludes our paper. 

\section{Related Work}
\label{sec:rel_work}
This section discusses other reviews on automated process extraction,
and outlines the contributions of our systematic review to the literature.
We differentiate between non-systematic and systematic literature reviews.

As a non-systematic review, Bellan et al. \cite{bellan2020qualitative} conducted a qualitative analysis in 2020 on process extraction methods. 
The authors highlight multiple limitations when applied to \textit{real-world} natural language text, 
and conclude that process extraction is a task that is far from solved. 
The same authors performed a non-systematic comparative analysis of the literature a year later~\cite{bellan2021process}, covering 13 papers published up to 2020, and categorised the used NLP techniques, supported process elements, and evaluations. 

We identified systematic reviews on process extraction up to 2018. 
These reviews thus do not cover important DL advancements that took place after 2018, notably BERT (i.e., Transformers)~\cite{devlin_bert_2018}.
The first relevant systematic review was conducted in 2016 by Riefer et al.~\cite{riefer2016mining}, reviewing 5 papers in total. 
The authors analyzed 3 aspects, namely textual input, text analysis, and model generation. 
Bordignon et al.~\cite{bordignon2018natural} performed a systematic review of works up to 2016 that apply NLP to the first 3 BPM lifecycle phases (i.e., identification, discovery and analysis), and categorised the NLP tools utilised.  
Maqbool et al.~\cite{maqbool2019comprehensive} covers process extraction up to 2018, focusing in particular on the extraction of Business Process Model and Notation (BPMN)~\cite{omg_bpmn} process models, categorising the used NLP tools and techniques, and BPMN modelling constructs and tools.
Schüler et al.~\cite{schuler_state_2024} performed a more general systematic review in 2022 (published in 2024) on the automated generation of process models from a range of input sources, including source code, business rules, event logs and natural text, among others.
The authors only shortly discuss process extraction (2 paragraphs).

Compared to these studies, we present a systematic review of NLP-enabled process extraction literature 
up to 2023:
we thus also include novel DL methods in our review. Further, we evaluate a more comprehensive set of review categories, including natural language analysis (text input, computational paradigm and tools), process generation (computational paradigm, intermediary representations, and target model), and evaluation (methods and dataset).
Finally, our review pays special attention to works that rely on innovations in ML/DL-based NLP for process extraction.

\section{Methods}
\label{sec:methods}
We performed a systematic review of the literature based on the PRISMA (Preferred Reporting Items for Systematic Reviews and Meta-Analyses) guidelines~\cite{Page2021}, 
which are widely recognised recommendations for conducting systematic reviews and meta-analyses.  
We searched 5 major scientific literature databases: Web of Science, Scopus, IEEE Xplore, ACM, and ScienceDirect, thus covering all disciplines relevant to our search.
Our search queries focused on the intersection of two key concepts: NLP and BPM\footnote{Any work focusing on the use of ML/DL for these concepts would thus also be included.}. 
Table~\ref{table:synonyms_search_queries} shows the synonyms and subfields (e.g., decision model extraction) considered for each concept and the corresponding search queries.
We considered the title, abstract, and author's keywords metadata fields in our search.


\begin{table}
\caption{Synonyms and search queries for NLP and BPM.}
\label{table:synonyms_search_queries}
\def\arraystretch{1.1}
{\small
\begin{tabular}{|p{2.5cm}|p{4.5cm}|p{6.25cm}|}
\hline
\textbf{Concepts} & \textbf{Synonyms} & \textbf{Search Queries} \\
\hlineB{2.5}

\multirow{2}{2.5cm}{Natural Language Processing} & NLP & ``NLP'' OR \\
\cline{2-2}
& Natural Language Processing & ``Natural Language Processing'' OR  \\
\cline{2-2}
& Text Mining & "Text mining'' OR \\
\cline{2-2}
& Text Analysis & ``Text analysis'' OR \\ 
\cline{2-2}
& NL Processing & ``NL processing'' OR \\
\cline{2-2}
& NLTK &  ``NLTK'' \\
\hline
\multirow{3}{2.5cm}{Business Process Management} & Process Modelling & ``Process Modelling'' OR \\
\cline{2-2}
& Business Process \newline Management &  ``Business Process Management'' OR \\
\cline{2-2}
& BPM & ``BPM'' OR  \\
\cline{2-2}
& Business Process Model and \newline Notation & ``Business Process Model and Notation'' OR \\
\cline{2-2}
& BPMN & ``BPMN'' OR \\
\cline{2-2}
& Process Extraction & ``Process Extraction'' OR \\
\cline{2-2}
& Process Management & ``Process Management'' OR  \\
\cline{2-2}
& Business Process Elicitation & ``Business Process Elicitation'' OR  \\
\cline{2-2}
& Decision Model and Notation & ``Decision Model and Notation''  OR \\
\cline{2-2}
& DMN &  ``DMN'' OR \\
\cline{2-2}
& Process Discovery & ``Process Discovery'' OR  \\
\cline{2-2}
& Decision Model Extraction &  ``Decision Model Extraction'' OR  \\
\cline{2-2}
& Business decision \newline management & ``Business decision management'' OR \\
\cline{2-2}
& Decision Extraction & ``Decision Extraction'' \\
\hline
\end{tabular}}
\end{table}


We removed duplicates from the search results and downloaded the remaining papers. The papers' titles, abstracts and keywords were screened for eligibility based on the inclusion/exclusion criteria described in the subsections below. In case a paper was still considered inconclusive after screening, it was fully reviewed to determine its  eligibility. 
We used the Covidence\footnote{\url{https://www.covidence.org/}} software to keep track of 
the review process. 

\subsection{Inclusion Criteria (IC)}
Articles meeting all of the following criteria were included in our systematic review: 
(\textbf{IC.1}) full research articles published in a peer-reviewed conference or journal, where a full text was available, and written in English (there was no restriction on time period); 
(\textbf{IC.2}) primary research articles, i.e., presenting original research contributions\footnote{In case of multiple papers related to the same contribution, we reviewed the most recent paper.}; (\textbf{IC.3}) articles specifically covering the use of NLP for process extraction from natural language text; (\textbf{IC.4}) articles that explicate a concrete method and describe experiments to empirically validate their method.

\subsection{Exclusion Criteria (EC)}
We excluded articles from our systematic review that did not comply with the IC as follows: (\textbf{EC.1})
articles that are not full research papers (such as posters, short papers, abstracts, reports, theses, or book chapters), not published in a conference or journal (e.g., workshops, discussion forums, white papers, technical reports), not peer-reviewed (including preprints), where no full text was available, or that were not written in English; (\textbf{EC.2}) secondary research articles, i.e., using primary research to derive results (e.g., literature reviews, meta-analysis, comments); (\textbf{EC.3}) articles not specifically covering the use of NLP for process extraction. This includes studies on the use of NLP for process redesign, matching or process prediction, sentiment analysis, works that target individual labels instead of natural text, or studies on generating natural text from processes; (\textbf{EC.4}) articles that only superficially discuss an applied method, or do not describe experiments to validate their method.

\subsection{Study Classification}
We classified the remaining studies according to the orthogonal themes in Table~\ref{tab:themes_categories}.
In line with our research questions, themes include NLP, process extraction, evaluation datasets and methods, and metadata to analyze publication trends. 
Based on the reviewed papers, we identified a set of categories per theme; in turn, these were used to classify the papers (i.e., qualitative inductive coding~\cite{Fereday2006}).

\begin{table}
\caption{Themes and categories used for classification.}
\label{tab:themes_categories}
\def\arraystretch{1.1}
{\small
\begin{tabular}{|m{1.8cm}|m{1.9cm}|m{5cm}|m{4.1cm}|}
\hline
\textbf{Theme} & \textbf{Category} & \textbf{Description} & \textbf{Possible Values} \\
\hlineB{2.5}
Natural \newline Language \newline Analysis & Text Input             & Restrictions on textual data.                                                                  & User/group stories, unrestricted or not mentioned.                    \\ 
\cline{2-4}
                          &  Computing paradigm       & NLP Computing paradigm. & Rules or ML/DL modules, end-to-end DL. \\
\cline{2-4}
                          & NLP Tools                   & Off-the-shelf NLP tools used.                                                                           & E.g., NLTK, spaCy, WordNet, CoreNLP.                                 \\ 
\hline
Process Model \newline Generation  & Computing Paradigm          & Computing paradigm for process generation.                                  & Knowledge-based, ML/DL.                       \\ 
\cline{2-4}
                          & Intermediary represent. & Use of notation-agnostic, intermediary representations.        &  E.g., graph-based, dependency tuples.  \\ 
\cline{2-4}
                          & Target \newline model       & Type of process model and notation that was targeted.                                                       & Control flow (imperative / declarative) vs. decisional. E.g., BPMN, DCL, DMN.                       \\ 
\hline
Evaluation                & Methods \& Metrics          & Methods and metrics used for evaluation.                                                                          & Methods: component-based / holistic; systematic / expert-based. Metrics: e.g., P, R, F1                          \\ 
\cline{2-4}
                          & Evaluation Dataset          & Dataset used for evaluation.     & 
                          Type (real, synthetic), target model, size, domain, availability, link.                                   \\ 
\hline
Metadata          & Year            & Year of Publication.                                                                                        & From 2011 to 2023.                                                  \\ 
\hline
\end{tabular}}
\end{table}


\section{Results}
\label{sec:results}
\begin{figure}[]
    \centering
    \includegraphics[width=12cm]{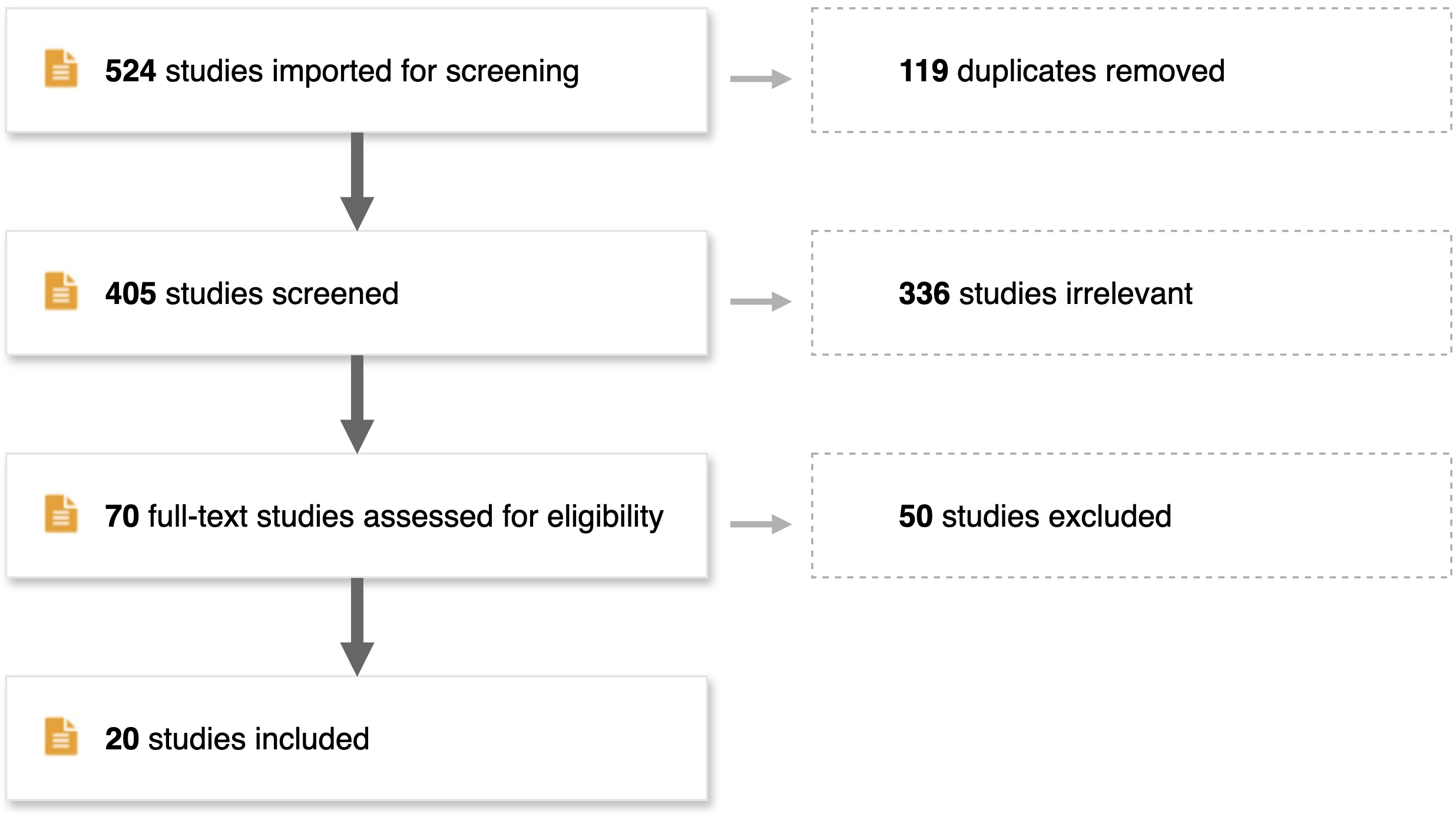}
    \caption{PRISMA flow diagram with results for our systematic review.}
    \label{fig:prisma}
\end{figure}

Figure~\ref{fig:prisma} shows the results of the systematic review process using the PRISMA flowchart. 
A total number of 524 studies were identified by our database search, 405 of which were unique. Table~\ref{tab:database_breakdown} shows the breakdown of search results per database.
Following title and abstract screening, 70 articles were retained for full-text screening. In the end, 20 articles were found to fully meet our inclusion/exclusion criteria.
Our search queries were executed in June 2023 and the found papers span a time period from 2011 to 2023.

\begin{table}[H]
\centering
\caption{Breakdown of search results per database.}
\label{tab:database_breakdown}
\def\arraystretch{1.1}
{\small
\begin{tabular}{|cc|c|}
\hline
\multicolumn{1}{|c|}{\textbf{Database}}            & \textbf{Web Address}                    & \textbf{Number of Results} \\ \hlineB{2.5}
\multicolumn{1}{|c|}{Scopus}              & https://www.scopus.com/        & 303              \\ \hline
\multicolumn{1}{|c|}{Web of Science}      & https://www.webofknowledge.com/ & 90               \\ \hline
\multicolumn{1}{|c|}{IEEE Xplore}         & http://www.ieeexplore.ieee.org & 33               \\ \hline
\multicolumn{1}{|c|}{ScienceDirect}       & https://www.sciencedirect.com  & 85               \\ \hline
\multicolumn{1}{|c|}{ACM Digital Library} & http://dl.acm.org              & 13               \\ \hline
\multicolumn{2}{|c|}{Total with Duplicates}                             & 524              \\ \hline
\multicolumn{2}{|c|}{Number of Duplicates}                                 & 119              \\ \hline
\multicolumn{2}{|c|}{Total without Duplicates}                              & \textbf{405}     \\ \hline
\end{tabular}
}
\end{table}

In the following sections, we answer our research questions RQ1-3 by discussing the reviewed papers in terms of the themes and categories from Table~\ref{tab:themes_categories}.
Appendix A includes tables that categorize each paper along these themes in detail.

\section{Natural Language Processing}
\label{sec:results_nlp}

This section discusses the NLP aspect of automated process generation, including restrictions on text input (Section~\ref{sec:nlp_text_input}), computational paradigms used to implement NLP (Section~\ref{sec:nlp_computing}), and publicly available NLP tools that were used (Section~\ref{sec:nlp_tools}).

\subsection{Text Input}
\label{sec:nlp_text_input}
In agile software development, user stories are used as a natural language medium to describe software requirements~\cite{nasiri2023automatic}. 
Nasiri et al.~\cite{nasiri2023automatic} extracted Unified Modeling Language (UML) activity diagrams from user stories
to capture software requirements as process models. 
In an information systems context, Goncalves et al.~\cite{goncalves2011let} similarly focus on user stories for process extraction. User stories, gathered through group storytelling, are described as a natural way for people to outline their everyday activities, difficulties, and suggestions for solving problems. 
The authors note that writing stories in a free style poses obstacles for NLP, such as ambiguity and lack of clarity; hence, the authors restrict the textual input to a scenario structure.

Halioui et al.~\cite{halioui2018bioinformatic} extract bio-informatics workflows from scientific texts.

The remainder of the papers did not explicitly mention a restriction on text input. 
That said, approaches that target decisional models will clearly expect text input with decision-related information; idem for control flow models.
Further, authors will evaluate their approach on datasets that, for instance, pertain to only one domain, or include sentences with minimal structural variability.
Hence, a method's (implicit) restrictions on text input may be reflected by the dataset used for evaluation.
We discuss evaluation datasets in Section~\ref{sec:eval_datasets}.


\subsection{Computing Paradigm}
\label{sec:nlp_computing}
Most reviewed papers involve a traditional ``NLP pipeline'', which identifies and tags textual elements with increasing amounts of semantics---ranging from part-of-speech and grammar relations to high-level semantics concepts---until the output is sufficiently detailed to generate a process model.
Although many customizations exist, a ``traditional NLP pipeline'' includes the tasks listed in Table~\ref{tab:nlp_tasks}.

\begin{table}[H]
  \centering  
  \small
  \caption{Traditional NLP pipeline tasks.}
  \label{tab:nlp_tasks} 
  \def\arraystretch{1.1}
  \begin{tabular}{|p{0.15\textwidth}|p{0.8\textwidth}|}    \hline
    \textbf{NLP Tasks} & \textbf{Description} \\    \hlineB{2.5}
        Tokenization & Splitting the raw text into chunks of words or sentences called \textit{tokens} (e.g., based on whitespaces, periods, or regular expressions). \\
        \hline
        Stemming and Lemmatizing & Stemming reduces words to their \textit{root form} by removing leading or trailing characters. Lemmatizing considers word meaning \& context to more accurately find root forms.\\
        \hline 
        Part-of-Speech (POS) Tagging & Assigning \textit{grammatical categories} (e.g., nouns, verbs) to words.\\ 
        \hline 
        Dependency \newline Parsing & Analyzing grammatical relationships between words in a sentence to reveal \textit{typed dependencies} (e.g., subject noun, verb object)~\cite{de2008stanford}.
        \\ 
        \hline 
        Named Entity Recognition (NER) & Identifying \textit{named entities} in text, such as names of persons, locations, and organizations. 
        This term is often used interchangeably with Entity Extraction (EE) and Entity Recognition (ER). \\ 
        \hline
        Coreference \newline Resolution & Linking pronouns (e.g., ``they'') and noun phrases (e.g., ``dog owner'') to the entities they refer to in the text (e.g., ``Bob'').
        This term is often used interchangeably with Anaphora Resolution and Entity Resolution.\\
        \hline 
        Semantic Analysis & Extracting and understanding the \textit{meaning of words or phrases} within text (e.g., activities, actors). In this category, we include \underline{Relation Extraction (RE)}, i.e., identifying semantic relations found in the text (e.g., sequence, parallel relations);
        \underline{Word Sense Disambiguation (WSD)}, i.e., disambiguating word meanings based on context; and \underline{word/sentence classification}, i.e., providing meaning to a word or sentence by assigning it a class.
        \\
        \hline 
        Preprocessing & NLP pipelines that leverage ML and DL models often require pre-processing by generating word embeddings (e.g., using GloVe~\cite{pennington2014glove}) or generating feature-vectors (e.g., using bag-of-words or TF-IDF).
        \\
        \hline
    \end{tabular} 
    
\end{table}

Other works rely on \textit{``non-traditional'' DL pipelines}, as we discuss later.

To answer \textbf{RQ2}, we proceed by summarising the use of ML/DL models for NLP tasks from Table~\ref{tab:nlp_tasks}. Subsequently, we discuss performance comparisons between different computing paradigms;
and describe non-traditional DL pipelines.

\begin{figure}[ht!]
    \centering
    \includegraphics[width=12cm]{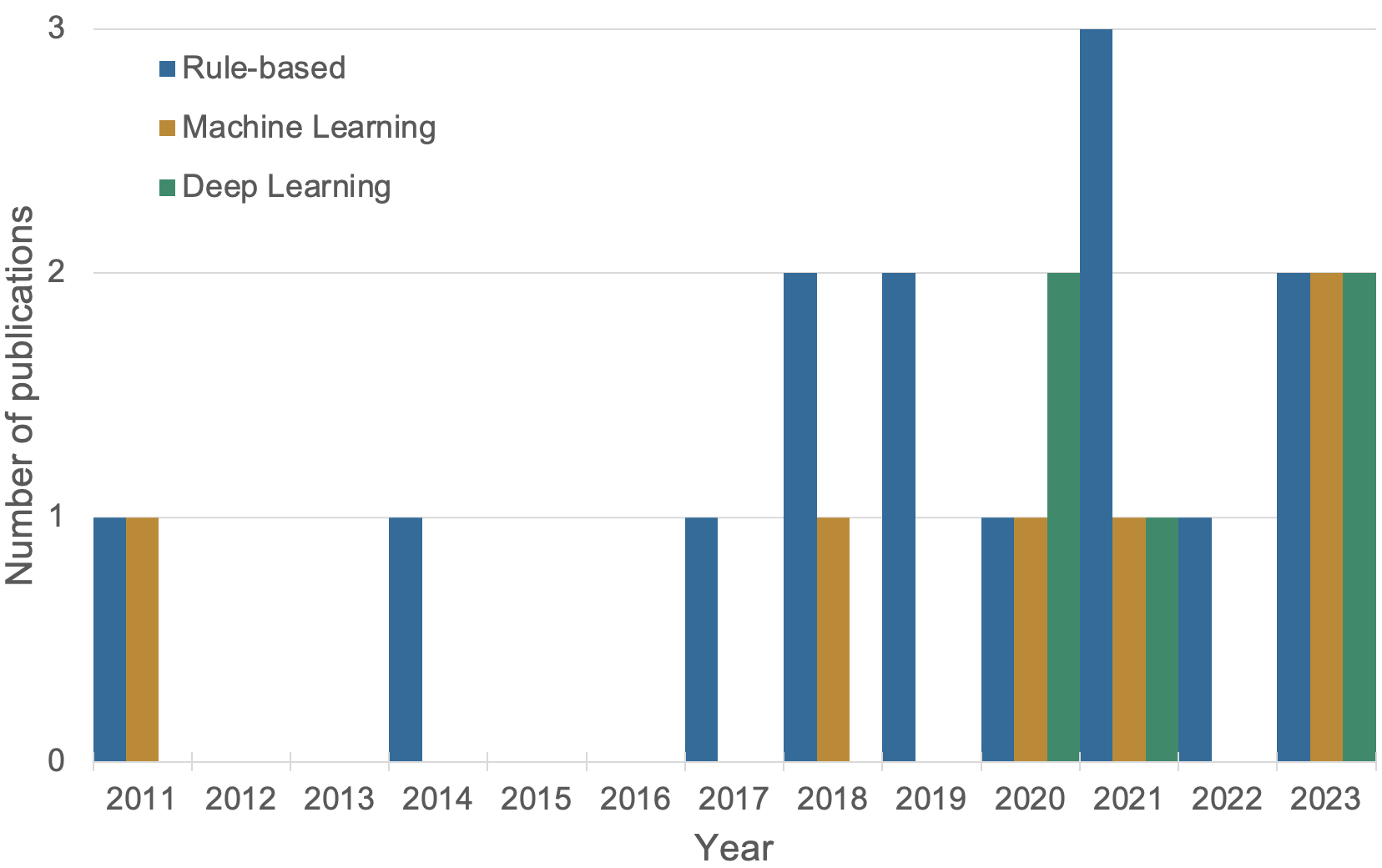}
    \caption{Papers published over time with their computational paradigms for NLP\protect\footnotemark.}
    \label{fig:nlp_method_year}
\end{figure}

Figure~\ref{fig:nlp_method_year} shows the evolution of computing paradigms used for NLP in reviewed papers from 2011-2023. 
In line with the advent of the DL BERT model in 2018, the first reviewed papers on DL-based process extraction appeared in 2020. 
In 2023, we find an even distribution across computing paradigms.

\footnotetext{Note that papers covering multiple paradigms will be counted once for each paradigm.}

Table~\ref{tab:ml-dl} summarises the ML/DL models used to implement NLP tasks.
\\

\begin{table}[H]
\caption{ML and DL models used for NLP tasks.}
\label{tab:ml-dl}
\def\arraystretch{1.1}
\small{
\begin{tabular}{|p{3.2cm}|p{10cm}|}
\hline
\textbf{Task} & \textbf{ML / DL Models} \\ 
\hlineB{2.5}                                                     
Preprocessing                & BERT (or possibly Word2Vec) for word embeddings~\cite{qian2020approach}; \newline Bag Of Words, TF-IDF, word embedding (GloVe)~\cite{goossens2023extracting};\newline RNN and Bi-LSTM for sentence- \& process-level encodings, ON-LSTM with process-level LM objective~\cite{han2020bps} \\ \hline
NER                          & Conditional Random Fields (CRF)~\cite{neuberger2023beyond}; \newline Pre-Trained BERT~\cite{goossens2023extracting,lopez_declarative_process_discovery} and Bi-LSTM~\cite{goossens2023extracting};                                                                                                                                                            \\ \hline
Coreference \newline Resolution       & Pre-trained DL model~\cite{neuberger2023beyond}; NeuralCoref4 
\cite{etikala2020text2dec,etikala2021extracting,goossens2023extracting} \\ \hline
\textit{Semantic Analysis}            &   \\                                                        
\hspace{0.2cm}RE                           & Catboost Gradient Boosting~\cite{neuberger2023beyond}; pre-trained BERT  (binary, multi-class)~\cite{lopez_declarative_process_discovery}
\\
\hspace{0.2cm}WSD                          & PAUM (Perceptron Algorithm with Uneven Margins)~\cite{halioui2018bioinformatic} \\
\hspace{0.2cm}Sentence/word \newline classification & Pre-Trained BERT (DistilBERT), logistic regression, na\"ive bayes and support vector machines~\cite{goossens2023extracting}; Bi-LSTM, CNN, MLP~\cite{qian2020approach}; Pre-trained BERT~\cite{lopez_declarative_process_discovery}\\ \hline    

\end{tabular}
}
\end{table}


\noindent \textbf{Comparison between rule-, ML- and DL-based NLP methods}.\\
Neuberger et al.~\cite{neuberger2023beyond} compared different NLP modules and pipelines:
\begin{enumerate}
\item An ML-based Relation Extraction (RE) module outperformed a rule-based RE module. The authors also noted that, while rule-based methods can be optimised for specific domains, they are hard to adapt to other applications.
\item Their NLP pipeline outperformed Jerex~\cite{Eberts2021}, an end-to-end DL method. The authors hypothesised that the used PET dataset~\cite{bellan2022process} is not yet extensive enough for training state-of-the-art DL models.
\end{enumerate}
Notably, their choice for Jerex was informed by the need for down-stream process generation: since Jerex's is also able to identify the textual location of an entity, its surrounding text can be used to obtain richer BPMN activity labels.

Goossens et al.~\cite{goossens2023extracting} make the following observations:
\begin{enumerate}
\item For sentence classification into dependency vs. logic sentences, a pre-trained BERT model (DistilBERT~\cite{sanh2020distilbert}) outperformed non-DL ML models (incl. logistic regression, na\"ive bayes and SVM).
\item For NER\footnote{In particular, extracting dependencies in terms of base, derived, and action tags.}, a pre-trained BERT model outperformed a trained-from-scratch Bi-LSTM-CRF model, 
although there were indications that the latter may work better on small datasets~\cite{ezencan2020comparison}.
\end{enumerate}

The authors also point out that the non-BERT models required extra pre-processing, respectively Bag of Words and TF-IDF, and word embeddings (GloVE~\cite{pennington2014glove}).

L\'opez et al.~\cite{lopez_declarative_process_discovery} compare ML-only, rule-only, and integrated ML + rule-based NER on roles, activities, and relations. 
The authors found the following:
\begin{enumerate}
\item Using only ML for NER yielded higher recall but lower precision;
\item Using only rules for NER yielded higher precision but lower recall;
\item Using ML for NER on roles and relations, and an integrated ML + rules for NER on activities, yielded the best F1 score.
\end{enumerate}

\noindent \textbf{Non-traditional, DL-based NLP pipelines}.\\
Qian et al.~\cite{qian2020approach} implemented the NLP aspect of process extraction using multi-grained text classification. In a course-grained phase, based on sentence encodings and features, sentences are classified (action vs. statement) and semantically tagged (e.g., sequential, concurrent). 
Next, based on the sentence-level encodings and features, together with word-level embeddings, a fine-grained phase semantically tags words and phrases with roles.
The authors use a combination of Bi-LSTM, CNN, and multiple MLPs (Multi-Layer Perceptrons). 
Notably, the authors use LSTM as it was deemed suitable for the sequential nature of text.
Han et al.~\cite{han2020bps} use RNN and Bi-LSTM to obtain sentence-level and process-level encodings; these are later passed to an Ordered Neurons LSTM (ON-LSTM)~\cite{shen2018ordered} for process model generation (Section~\ref{sec:proc_extract_computing}).

In addition to a traditional NLP pipeline, Neuberger et al.~\cite{neuberger2023beyond} use an end-to-end DL model (Jerex~\cite{Eberts2021}) to perform RE and NER.

Halioui et al.~\cite{halioui2018bioinformatic} combine ML with an ontology for semantic analysis. 
The authors map text elements to concepts from this ontology; precedence constraints from the ontology are then used to help reconstruct the process. 
The authors further consider the word's context to perform WSD; i.e., POS tags and mapped concepts surrounding the word are used as context to train a ML (PAUM~\cite{li2005using}) model. 
E.g., this allows mapping the term ``MEGA'' to different concepts based on surrounding context. 
Notably, PAUM was designed for imbalanced data and has successfully been applied to semantic annotation.

\subsection{NLP Tools}
\label{sec:nlp_tools}

Table~\ref{table:nlp_tools} lists the NLP tools found in the reviewed papers.
Many tools used in the reviewed papers have overlapping capabilities such as lemmatization, dependency parsing, POS tagging, and coreference resolution. 
Tools like Stanford POS Tagging, Stanford Parser, and spaCy can handle multiple tasks, from dependency tree generation to semantic role labeling.
Specialised tools such as NeuralCoref4 or Ansj Tokeniser target specific tasks, i.e., coreference resolution and word segmentation.  
Domain-specific tools such as MedPos focuses on POS tagging of clinical text.

\begin{table}[H]
\caption{Public NLP tools used for process extraction.}
\label{table:nlp_tools}
\def\arraystretch{1.1}
{\small 
\begin{tabular}{|p{2.8cm}|p{8.9cm}|p{1.5cm}|}
\hline
\textbf{NLP Tools} & \textbf{Description} & \textbf{Papers} \\ 
\hlineB{2.5}

Ansj Tokeniser & Chinese word segmentation tool. & \cite{chen2014pewp}\\
\hline
ELMo & A pre-trained model that generates embeddings for words based on their context across linguistic contexts. & \cite{qian2020approach}\\
\hline
FrameNet & A lexical database that annotates words in terms of their semantic frames and roles. & \cite{sintoris2017extracting}\\
\hline
GloVe & An unsupervised learning algorithm for obtaining word vector representations based on word-word co-occurrence. & \cite{qian2020approach,han2020bps}\\
\hline
MedPos Tagging & Assigns POS tags to words targeting \textit{clinical text}. & \cite{halioui2018bioinformatic}\\
\hline
NeuralCoref4 & A spaCy extension that annotates and resolves coreference clusters using a neural network.  & \cite{etikala2020text2dec,etikala2021extracting,goossens2023extracting}\\
\hline
NLTK & Library with tools for symbolic and statistical NLP. & \cite{goncalves2011let,etikala2020text2dec}\\
\hline
OpenNLP \newline Chunker & Identifies syntactic constituents (e.g., noun, verb phrases). & \cite{halioui2018bioinformatic}\\
\hline
spaCy & Python NLP library. & \cite{etikala2021extracting,etikala2020text2dec,ferreira2017semi,honkisz2018concept}\\
\hline
Stanford CoreNLP & Java suite of NLP tools. & \cite{etikala2020text2dec,nasiri2023automatic}\\
\hline
Stanford \newline Lemmatiser & Converts words to their root form. & \cite{sonbol2023machine,halioui2018bioinformatic}\\
\hline
Stanford Parser & Parses sentences to derive their grammatical structure. & \cite{friedrich2011process,sintoris2017extracting, honkisz2018concept}\\
\hline
Stanford POS \newline Tagging & Assigns POS tags to words in an input text. & \cite{etikala2021extracting,sonbol2023machine,vda_extracting_declarative_process}\\
\hline
Stanford Tregex & Queries syntactic dependency trees based on patterns. & \cite{quishpi2021extracting}\\
\hline
Stanza & Python NLP library developed by Stanford for multiple human languages, similar in functionality to CoreNLP. & \cite{sholiq2022generating}\\
\hline
Word2Vec & A technique that employs neural networks to learn word embeddings from a text corpus. & \cite{qian2020approach}\\
\hline
WordNet & A lexical database of English, grouping words into sets of synonyms and providing brief definitions. & \cite{honkisz2018concept,sintoris2017extracting,friedrich2011process,sonbol2023machine,lopez_assisted_declarative_process,vda_extracting_declarative_process}\\
\hline

\end{tabular}}
\end{table}

\section{Process Model Generation}
\label{sec:results_proc_extract}

This section discusses the generation of process models from NLP output: 
including the computational paradigm used (Section~\ref{sec:proc_extract_computing}), intermediary representations, if any (Section~\ref{sec:proc_extract_repr}), and targeted type of process model (Section~\ref{sec:proc_extract_target}).

\subsection{Computing Paradigm}
\label{sec:proc_extract_computing}

We observe that, despite the increasing popularity of ML/DL methods in NLP pipelines (Section~\ref{sec:results_nlp}), the majority of studies rely on knowledge-based methods for process generation---with the caveat that ML/DL enabled NLP pipelines may directly output a process model (see ``Machine Learning / Deep Learning'' below). 
We discuss these two approaches below.\\
[0.2cm]
\textbf{Knowledge-based Methods}.\\
Here, we consider the use of any knowledge-based artefact, such as concrete rules (e.g., Prolog), templates, or custom algorithms, to generate process elements based on NLP output. 
We discuss the type of artefact and the NLP output involved: 

\begin{itemize}
\item \textit{Concrete Rules}. Nasiri et al.~\cite{nasiri2023automatic} provide a set of Prolog rules, which refer to Part-of-Speech (POS) tags and typed dependencies from NLP output, in order to generate activity diagrams.
Halioui et al.~\cite{halioui2018bioinformatic} use JAPE (Java Annotation Patterns Engine)~\cite{cunningham2013getting} pattern-based rules to, among other things, avoid incorrect entity recognitions. For instance, in case a noun, verb or adjective is in lowercase, a rule will ensure it is not matched to a software program\footnote{This could be considered part of NLP; but, we mention it here as they act on NLP output.}.

\item \textit{Custom algorithms}. 
Sonbol et al.~\cite{sonbol2023machine} present a custom algorithm to generate an intermediary Text Graph (i.e., process model) based on semantic concepts, their tags and relations from the NLP step.
Chen et al.~\cite{chen2014pewp} collect a ``verb flow'' from a given start word using a recursive algorithm, based on POS tags, grammar relations, and semantic tags (topics). E.g., for topic ``House Cleaning'', this process would yield Clean→Dust Collection→Scrub→Wash→Wipe out.
Van der Aa et al.~\cite{vda_extracting_declarative_process} define an algorithm based on elements from NLP output that indicate modality (e.g., ``must''), negation, connectors (e.g., ``or'') and temporal relations (e.g., ``before''); and consider verb-based (e.g., ``create'') and noun-based (e.g., ``creation'') activities. 
E.g., a sequential relation \textit{rel(A,B)}, where activity A is mandatory but B is not, results in a \textit{precedence} constraint. 
L\'opez et al.~\cite{lopez_declarative_process_discovery} apply an algorithm to integrate the results from RE and NER; when the binary RE classifier detects a sentence with a relation, the NER model extracts 2 events, with a relation as detected by the multi-class RE classifier.

\item \textit{Templates / patterns}. 
Etikala et al.~\cite{etikala2020text2dec} apply predefined patterns, such as ``passive A\textless=B'' (A follows from B) and relevant verbs (e.g., ``determine'') to generate dependency tuples in the form of \textlangle action, base, derived\textrangle~concepts (e.g., determine, height, Body Mass Index), based on typed dependencies and semantic concepts from the text. Later on, Etikala~\cite{etikala2021extracting} additionally extracts decision logic by applying patterns with markers such as ``if”, ``then”, etc., which is ultimately converted into decision tables.
Ferreira et al.~\cite{ferreira2017semi} define high-level templates to identify BPMN elements (e.g., XOR gateway) based on POS tags and semantic tags (e.g., condition, conjunction).
Quishpi et al.~\cite{quishpi2021extracting} use a query language (Tregex) to capture decision requirements (e.g., determine obesity from height) from a syntactic dependency tree with typed dependencies. 

\item \textit{High-level Rules}. These include ``if-then'' rules narratively described in tables or text. 
(For instance, these could be implemented using concrete rules or a custom algorithm; see above.)
Sholiq et al.~\cite{sholiq2022generating} describe rules that, based on typed dependencies, extract ``fact types'' (FT); e.g., a binary FT would be ``officer \textit{accepts} document". These fact types are mapped to BPMN elements.
Goncalves et al.~\cite{goncalves2011let} describe rules to find activities, actors and other elements, based on POS tags and typed dependencies.

\item \textit{Custom}. 
Others narratively describe a series of steps in the paper text. 
Honkisz et al.~\cite{honkisz2018concept} describe steps for identifying participants, subject-verb-object and BPMN gateway keywords, based on typed dependencies and semantic concepts from the text. Azevedo et al.~\cite{azevedo2018bpmn} only shortly summarise their process generation approach; based on NLP output (e.g., POS tags), the authors generate process elements based on the sequential ordering of sentences. 
 Friedrich et al.~\cite{friedrich2011process} describe an extensive process that includes extracting actors, actions and objects from text; merging actions, based on anaphora resolution; and generating process elements (e.g., exclusive, parallel) based on textual markers (e.g., ``or'', ``in parallel'').
L\'opez et al.~\cite{lopez_assisted_declarative_process} suggest process elements from natural text for confirmation by the user, based on synonyms from Wordnet and a domain-specific R-A-R (role-activity-relation) dataset that is updated based on user interactions. 
So-called \textit{highlights} keep the link between text and associated process elements.

\end{itemize}

\noindent \textbf{Machine Learning / Deep Learning}.\\
To the best of our knowledge, ML/DL is only used once to implement process generation in the reviewed papers~\cite{han2020bps}.
Some authors cast process extraction as NLP problems (e.g., text classification, NER), which they then solve using ML/DL methods~\cite{qian2020approach,goossens2023extracting}. Here, NLP thus directly yields the targeted output; 
however, extra work may still be needed to generate a complete process model from this output.

Qian et al.~\cite{qian2020approach} cast process extraction as a multi-grained text classification problem; output includes semantic annotations of sentences and words/phrases. 
Goossens et al.~\cite{goossens2023extracting} use the results from NER, i.e., a set of dependency tags (action, base, derived concepts) to generate dependency tuples; a tuple consists of dependency tags from 1 sentence. (Etikala et al.~\cite{etikala2020text2dec} apply a separate knowledge-based step for extracting dependency tuples from NLP output.) 
Neuberger et al.~\cite{neuberger2023beyond} focus on RE and NER for the purpose of process extraction; the latter is considered out of scope.

In contrast to these works, Han et al.~\cite{han2020bps} pass the NLP output, i.e., sentence-level and process-level encodings, to a novel ML model called Ordered Neurons LSTM (ON-LSTM)~\cite{shen2018ordered}, which separately implements process model generation. In a structure-retrieve step, the latent hierarchical process structure is inferred from the ON-LSTM model. A BPMN diagram can then be extracted using Jinja2\footnote{https://jinja.palletsprojects.com/en/3.1.x/}.

\subsection{Intermediary Representation}
\label{sec:proc_extract_repr}

Here, we include any method where a notation-agnostic, intermediary representation is separately generated. This representation can then be mapped to multiple different model notations.
Note that we do not consider intermediate NLP output (e.g., dependency trees, semantic tags) here, nor internal model representations (e.g., latent representations in DL models).

\begin{itemize}
\item \textit{Graph-based Model}.
Sonbol et al.~\cite{sonbol2023machine} construct a Text Graph with sentences as vertices, which are connected based on their ordering in the text (i.e., 2 consecutive sentences as 2 connected vertices). They also construct a concept map that includes interrelated actors and resources. This concept map is leveraged to further connect sentence vertices, i.e., based on their shared concepts in the map. The Text Graph is converted into a BPMN model in an 8-step procedure.
Friedrich et al.~\cite{friedrich2011process} generate a graph-based World Model that capture actors, resources, actions and flows as relations between them. Subsequently, a 9-step process generates a BPMN model from the World Model.
We note that Azevedo et al.~\cite{azevedo2018bpmn} mention an intermediary \textit{Tree-based} Process Model, i.e., with sentences as tree nodes, 
but do not provide details.

\item \textit{Table}. 
Honkisz et al.~\cite{honkisz2018concept} generate a table with columns including ordering, activity, condition and actor. The authors subsequently generate a BPMN diagram but do not detail the individual steps.

\item \textit{Dependency tuples}. Goossens et al.~\cite{goossens2023extracting} and Etikala et al.~\cite{etikala2020text2dec,etikala2021extracting} generate intermediary dependency tuples that are ultimately converted into a Dependency Requirements Diagram (DRD).

\end{itemize}


\subsection{Target Process Model}
\label{sec:proc_extract_target}

Regarding the type of process model that is targeted, we find a contrast between \textit{control flow models }(BPMN, UML activity diagrams) and \textit{decisional models} (Decision Model and Notation, DMN: decision requirements, tables). 
Some studies focus only on particular elements; e.g., Chen et al.~\cite{chen2014pewp} capture activity ordering (``verb-flow'').

In the control flow category, we further distinguish \textit{imperative models}, i.e., which fully specify the activity occurrence and sequential ordering; and \textit{declarative models}, i.e., which include high-level constraints on the occurrence and ordering of certain activities (e.g., precedence, response, succession~\cite{vda_extracting_declarative_process}). 

Table~\ref{table:target_model} shows a breakdown of the reviewed papers based on the type of target model and the particular notation targeted, if any.
Overall, imperative control flow models are the most covered (15), followed by decisional models (4) and then declarative control flow models (3).

\begin{table}[H]
\caption{Target Process Models}
\label{table:target_model}
\def\arraystretch{1.1}
{\small 
\begin{tabular}{|p{2.3cm}|p{6.5cm}|p{3.5cm}|}
\hline
\textbf{Model Type} & \textbf{Notation} & \textbf{Papers} \\
\hlineB{2.5}
Control flow: \newline Imperative & BPMN & \cite{sonbol2023machine,sholiq2022generating,azevedo2018bpmn,ferreira2017semi,friedrich2011process,sintoris2017extracting,bellan2022process,han2020bps,qian2020approach,honkisz2018concept,goncalves2011let}\\ 
& UML activity diagram~\cite{omg_uml} & \cite{nasiri2023automatic}\\
& Custom / unspecified & \cite{chen2014pewp,neuberger2023beyond,halioui2018bioinformatic}\\
\hline
Control flow: \newline Declarative & Declare~\cite{wvda_declare} & \cite{vda_extracting_declarative_process} \\
& DCR (Dynamic Condition Response)~\cite{hildebrandt_dcr} & \cite{lopez_assisted_declarative_process,lopez_declarative_process_discovery} \\
\hline
Decisional & DMN~\cite{omg_dmn} & \cite{etikala2021extracting,goossens2023extracting,etikala2020text2dec,quishpi2021extracting}\\
\hline
\end{tabular}}
\end{table}

\section{Process Extraction Evaluation}
\label{sec:results_eval}

In this section, we review the datasets and methods commonly utilised to evaluate automated process extraction.

\subsection{Evaluation Datasets}
\label{sec:eval_datasets}

Table~\ref{tab:eval_datasets} shows the datasets used by the reviewed papers. Regarding \textit{Type of Data}, \textit{synthetic} means that the included sentences were manually written by the authors; \textit{real} means that sentences were collected from real-world sources, such as academic papers, laws and regulations, interviews and Internet searches.
We note that the \textit{Dataset Size} column lists the level of detail provided by the authors; 
e.g., if the paper mentions X ``example texts'', we repeat that in this column (as this does not necessarily constitute X separate cases).
Below, we make several observations on these datasets; we also discuss multiple real datasets and how they were collected.
\\
[0.2cm]
\textbf{Main Observations on Evaluation Datasets}.
\begin{enumerate}
    \item There is a lack of consistently used evaluation datasets. 
Only the Friedrich~\cite{friedrich2010automated} and PET~\cite{bellan2022process} datasets are re-used, i.e., in 2 other studies (also, per dataset, only 1 of these 2 studies was not authored by its creators). 

    \item Many online datasets are no longer available. Some articles provide links to their evaluation dataset; however, Table~\ref{tab:eval_datasets} shows that 4/9 links no longer work\footnote{We suggest readers to reach out to the paper authors to gain access to these datasets.}.

    \item There is a paucity of gold-standard datasets. There is currently 1 reference dataset for control flow process extraction, i.e., the PET dataset (Section~\ref{sec:eval_datasets}).

    \item Most evaluation datasets are small in size. While a variety of datasets are used, they are relatively small in size (i.e., tens or hundreds of process descriptions). 
\end{enumerate}

\noindent \textbf{Real-world Datasets and Their Collection}.\\
Friedrich et al.~\cite{friedrich2010automated} introduced a dataset consisting of 47 text-model pairs, each pair including a textual description and a corresponding BPMN model created by a human modeler. It spans processes from academic sources, industry (e.g., BPM tool vendors), textbooks and the public sector. Hence, it can be considered a mix of synthetic (e.g., textbooks) and real (e.g., industry).
Bellan et al.~\cite{bellan2022process} introduced the PET dataset, which was constructed based on the above ``Friedrich dataset''. It improves upon the granularity of the latter by annotating parts of the text with corresponding process elements. 
Also, since the textual dataset was annotated by 3 annotators, it can be considered more of a gold standard.
The authors further provide baseline results for typical process extraction-related NLP tasks using the PET dataset, i.e., NER and RE using ML (CRF) and a custom rule-based approach.

\begin{table}[H]
\caption{Evaluation datasets used in process extraction.}
\def\arraystretch{1.1}
\label{tab:eval_datasets}
\centering
\small
\begin{tabular}{|p{1.1cm}|p{1.5cm}|p{4cm}|p{1.2cm}|p{4.5cm}|p{2.2cm}|}
\hline

\textbf{Type} & \textbf{Target Model} & \textbf{Dataset Size} & \textbf{Public} & \textbf{Domain} & \textbf{Notes \& \newline Papers} \\ 
\hlineB{2.5}
Synth./ Real & Control (Imper.) & 47 Processes, 432 Sentences & Public\textsuperscript{1} & Academic, industry, textbook, public sector & `Friedrich'~\cite{friedrich2010automated} \newline  \cite{friedrich2010automated,friedrich2011process,sonbol2023machine} \\
\hline
Synth./ Real & Control (Imper.) & 45 Processes, 417 Sentences & Public\textsuperscript{2} & Cfr. Friedrich et al.~\cite{friedrich2010automated} & `PET'~\cite{bellan2022process} \newline \cite{bellan2022process,bellan_pet_23,neuberger2023beyond} \\
\hline
Real & Control (Imper.) & Ca. 260 Processes & Private & Business processes (RPA, BluePrism, SAP) & \cite{han2020bps} \\
\hline
Synth. & Control (Imper.) & 10 Cases & Public\textsuperscript{3} & Finance and HR & \cite{sholiq2022generating}\\
\hline
Real & Control (Imper.) & 56 Texts & Private & US visa process, Federal Network Agency of Germany, others & \cite{ferreira2017semi} \\
\hline
Real & Control (Imper.) & - & Public\textsuperscript{4} & Internet searches; 5 topics (e.g., house cleaning) & \cite{chen2014pewp} \\
\hline
Synth. & Control (Imper.) & - & Public\textsuperscript{5} & Computer repair & \cite{honkisz2018concept} \\
\hline
Real & Control (Imper.) & Sentences: 2636(COR), 2172(MAM) & Public\textsuperscript{6} & Cooking, maintenance & \cite{qian2020approach} \\
\hline
Real & Control (Imper.) & 320 Texts & Public\textsuperscript{7} & Keywords: Phylogenetic, Data, InferenceProgram (PubMed PCM) & \cite{halioui2018bioinformatic} \\
\hline
Synth. & Control (Imper.) & 2 Cases (33, 17 Activities) & Public\textsuperscript{8} & Finance and Web & \cite{nasiri2023automatic} \\
\hline
Real & Control (Imper.) & 2 Cases (26, 15 Story Events) & Private & Course enrolment, manual process elicitation & \cite{goncalves2011let} \\
\hline
Synth./ Real & Control (Decl.) & 103 Constraint Descriptions & Public\textsuperscript{9} & Industrial, academic (general, declarative) & \cite{vda_extracting_declarative_process} \\
\hline
Synth./ Real & Control (Decl.) & 107 Process Descriptions & Public\textsuperscript{10} & BPM academic initiative, students, social services & \cite{lopez_declarative_process_discovery} \\
\hline
Synth. & Decis. & 2 Cases & Public\textsuperscript{11} & Finance and Health & \cite{etikala2020text2dec} \\
\hline
Synth. & Decis. & 20 Texts & Private & - & \cite{etikala2021extracting} \\
\hline
Real & Decis. & Sentences: 577 (train), 232 (test); Text: 6 (38-101 words) & Public\textsuperscript{12} & Sentences: papers, laws \& regulations, Etikala et al.~\cite{etikala2020text2dec}, others; Text: DM community, Internet & \cite{goossens2023extracting} \\
\hline
Synth. & Decis. & 12 Texts & Public\textsuperscript{13} & Finance, healthcare & \cite{quishpi2021extracting} \\
\hline
\end{tabular}

\end{table}

{\footnotesize
We could not find a description of the dataset in Azevedo et al.~\cite{azevedo2018bpmn} and L\'opez et al.~\cite{lopez_assisted_declarative_process}.

\begin{enumerate}
\item [1] \url{http://github.com/FabianFriedrich/Text2Process}
\item [2] \url{http://huggingface.co/datasets/patriziobellan/PET}
\item [3] Listed in paper
\item [4] \url{http://yun.baidu.com/share/link?shareid=3648609423&uk=1862989032}
(not working at submission)
\item [5] Listed in paper
\item [6] \url{https://github.com/qianc62/MGTC}
\item [7] \url{https://data.world/growler/datasets}
\item [8] Listed in paper
\item [9] \url{http://github.com/hanvanderaa/declareextraction}
\item [10] \url{http://github.com/RasmusIven/DCR-Textograph} (not working at submission)
\item [11] Listed in paper
\item [12] \url{http://github.com/Goossens496/Extracting-DMN-models-from-text} (not working at submission)
\item [13] \url{http://github.com/ProjectTex2Dec/Text2Dec/tree/master/data/collected} (not working at submission)
\end{enumerate}
}

Han et al.~\cite{han2020bps} collected 210 process descriptions from BluePrism RPA (Robotic Process Automation), and 50 process descriptions from the Internet such as whitepapers, user guides, and SAP support documents. 
Ferreira et al.~\cite{ferreira2017semi} 
evaluated their system using 56 natural language text from the US immigrant visa process, Federal Network Agency of Germany, and others. 
Chen et al.~\cite{chen2014pewp} used a search engine to retrieve process descriptions on 5 topics (e.g.,  house cleaning).

To evaluate multi-grained classification for process extraction, Qian et al.~\cite{qian2020approach} manually collected a dataset consisting of (a) Cooking Recipes (COR) with 200+ recipes from recipe.com; (b) Maintenance Manuals (MAM) with 160+ device maintenance descriptions from ifixit.com. 
Halioui et al.~\cite{halioui2018bioinformatic} conducted a search on the PubMed Central database for papers with ``phylogenetic'' and filtered on articles with suitable bio-informatics workflows (i.e., keywords ``Data'', ``InferenceProgram'').

Goncalves et al.~\cite{goncalves2011let} manually collected user stories (events, characters, and other elements) from people involved with course enrolment (26 story events) and managing process elicitation at a company (15 story events). 

For declarative process extraction, Van der Aa~\cite{vda_extracting_declarative_process} gathered constraint descriptions from industrial and academic sources, with both declarative and general processes. L\'opez et al.~\cite{lopez_declarative_process_discovery} gathered process descriptions from the BPM academic initiative, student interviews, and excerpts related to social services. 

For decisional model extraction, Goossens et al.~\cite{goossens2023extracting} collected a real-world dataset from the DM community, laws and regulations, Web searches, and their prior work~\cite{etikala2020text2dec}.
Quishpi et al.~\cite{quishpi2021extracting} collected 12 text-model pairs (i.e., text paired with the corresponding DMN model created by a human) based on materials from other authors. 

\subsection{Evaluation Methods}
\label{sec:eval_methods}

We observed 2 orthogonal sets of evaluation methods for process extraction:\\
[0.1cm]
\textbf{Component-based vs. Holistic Evaluation}

\textit{Component-based evaluation} evaluates the output of individual modules in the NLP and process model generation steps.
This has the advantage of locating specific pain points, and avoids errors from propagating to subsequent modules~\cite{neuberger2023beyond}.
It is typically combined with a holistic evaluation to evaluate overall pipeline performance.

\textit{Holistic evaluation} evaluates the holistic extracted process model; e.g., by comparing its individual elements 
with those of a reference model (systematic), or evaluating their understandability and usefulness
(expert-based). 
While this approach thus evaluates the overall pipeline performance, it may gloss over the contribution of (and issues with) individual modules.\\
[0.2cm]
\textbf{Systematic vs. Expert-based Evaluation}

\textit{Systematic comparison of generated and reference output}.
The generated output, albeit individual module output or the final process model,
is systematically compared to a reference model. 
This can be an \textit{element-by-element} comparison or based on an \textit{aggregate} measure, 
such as processes' behavioral profiles~\cite{weidlich2011}.
Here, \textit{graph-based} evaluations represent process models as graphs, and map the nodes between reference and extracted process graphs.
A potential drawback is that the reference model may be only one of many valid process models for the same text~\cite{sholiq2022generating}.

\textit{Expert evaluation of generated output}.
Experts manually evaluate the generated output, providing their opinions on metrics such as understandability, usefulness, their agreement with the output, or the model's equivalency with the text. Here, the evaluation can again be \textit{element-by-element} (i.e., evaluating individual elements), or focusing on properties of the output as a whole (\textit{aggregate}).
This avoids the above issue of multiple valid reference models, as it does not involve comparisons with a reference. 
However, it is also resource-intensive and relies on subjective judgments\footnote{Reference models (\textit{systematic}) will typically be created by experts, and thus also include a degree of subjectivity. The difference here is how the comparison is performed.}.

\vspace{0.2cm}
Table \ref{tab:eval_methods} shows a breakdown of the reviewed papers along these categories. 
Below, we separately discuss systematic and expert-based evaluations from the papers.\\
[0.5cm]
\textbf{Systematic Evaluations}\\
The majority of evaluation methods are \textit{holistic} and \textit{systematic (element-by-element)}: individual elements from the extracted and reference process model are systematically compared. 
Clearly, the evaluated elements will depend on the target process model (control flow vs. decisional). 
Instead of \textit{element-by-element}, Qian et al.~\cite{qian2020approach} utilise an \textit{aggregate} measure: assessing the similarity of the extracted and reference process model based on their behavioral profiles~\cite{weidlich2011}.

\begin{table}[H]
  \centering  
  \small
  \caption{Evaluation methods used in process extraction.}
  \label{tab:eval_methods} 
  \def\arraystretch{1.1}
  \begin{tabular}{|p{2.2cm}|p{2.2cm}|p{7.8cm}|p{1.2cm}|}    \hline
\textbf{Category 1}      & \textbf{Category 2}   & \textbf{Description}                                                                             & \textbf{Papers} \\ \hlineB{2.5}
Component-based & Systematic   & Components: Sentence classification                                                     & \cite{goossens2023extracting}                                           \\ \cline{3-4}
                &              & Components: Entity, Relation Extraction                        & \cite{bellan_pet_23,neuberger2023beyond}                                \\ \cline{3-4}
                &              & Components: Sentence classification, semantics recognition, semantic role labeling      & \cite{qian2020approach}                                                 \\ \hline
Holistic        & Systematic   & Elements: decisional dependencies \newline (dependency tuples and/or decision logic)                                                      & \cite{goossens2023extracting,etikala2020text2dec,etikala2021extracting,quishpi2021extracting} \\ \cline{3-4}
                &              & Elements: disambiguated words, extracted relations                           & \cite{halioui2018bioinformatic}                                         \\ \cline{3-4}
                &              & Elements: extracted entities, relations          & \cite{bellan_pet_23,neuberger2023beyond,lopez_assisted_declarative_process}                                \\ \cline{3-4}
                &              & Elements: activity sorting, found activities                                & \cite{chen2014pewp}                                                     \\ \cline{3-4}
                &              & Elements: UML activity/ BPMN / interm. repr. elements                                                 & \cite{ferreira2017semi,nasiri2023automatic,sholiq2022generating,honkisz2018concept}        \\ \cline{3-4}
                &              & Graph-based: Graph Edit Distance                                                        & \cite{friedrich2011process,sonbol2023machine}                           \\ \cline{3-4}
                &              & Graph-based: Similarity score & \cite{han2020bps}                                                       \\ \cline{3-4}
                &              & Aggregate measure: behavioral profiles~\cite{weidlich2011}              & \cite{qian2020approach}                                                 \\ \cline{3-4}
                &               & Elements: declarative constraints       & \cite{vda_extracting_declarative_process,lopez_declarative_process_discovery} \\ \cline{2-4}
                & Expert-based & Assess understandability and usefulness of extracted processes (aggregate)                         & \cite{sonbol2023machine}                                                \\ \cline{3-4}
                &              & Assess knowledge equivalency between extracted, reference processes (aggregate)                     & \cite{azevedo2018bpmn}                                                  \\ \cline{3-4}
                &              & Gather opinions (fully-agree to disagree) on extracted processes (aggregate)                        & \cite{ferreira2017semi}                                                 \\ \cline{3-4}
                &              & Separate valid (useful) from noise (not useful) extracted process elements (element-by-element)              & \cite{goncalves2011let}                                                 \\ \cline{3-4}
                &              & Manually compare textual elements with extracted elements (element-by-element)                               & \cite{ferreira2017semi} \\ \hline
\end{tabular}
\end{table}
Often in addition, several authors evaluate their approach in a \textit{component-based} and \textit{systematic (element-by-element)} way.
Etikala~\cite{etikala2021extracting} separately evaluates the decision logic and dependency tuple extractor components; individual output elements (e.g., dependency tuples) are compared to reference elements. 
Goossens et al.~\cite{goossens2023extracting} evaluate their sentence classification 
and downstream decisional model extraction. 
To avoid errors early on in the pipeline from propagating, 
Neuberger et al.~\cite{neuberger2023beyond} evaluate modules' performance in isolation by also providing reference inputs.
Qian et al.~\cite{qian2020approach} evaluate the accuracy of the classifiers involved in NLP and process extraction.

In \textit{graph-based} systematic evaluations, reference and extracted processes are represented as graphs. 
Then, based on a mapping of graph nodes between the reference and extracted process graph, authors calculate a similarity score.
Friedrich et al.~\cite{friedrich2011process} and Sonbol et al.~\cite{sonbol2023machine} rely on the Graph Edit Distance~\cite{dijkman2011}, which, based on the similarity of individual nodes,
calculates the number of node and edge edits needed to get from one graph to another.
Han et al.~\cite{han2020bps} similarly map identical nodes, in this case, based on their ``step'' descriptions and ``parent'' and ``previous'' nodes. A similarity score is calculated as the ratio of mapped nodes vs. the total number of model nodes.

Systematic evaluations typically evaluate correctness (false positives) and completeness (false negatives); hence, evaluation metrics tend to involve precision, recall, and F1-score. 
Sholiq et al.~\cite{sholiq2022generating}
calculate the Magnitude of Relative Error (MRE) as the ratio of differences between the extracted and reference model vs. the size of the reference model. 
Chen et al.~\cite{chen2014pewp}, in line with their focus on activity orderings (``verb-flow''), calculate 
the Normalised Distance-based Performance Measure (NDPM) as a ratio based on incorrectly \& correctly sorted and concurrent activities.
Van der Aa~\cite{vda_extracting_declarative_process} proposes a fine-grained ``slot filling'' approach, where the type of constraint and its arguments are ``slots''; precision and recall are based on the number of matching slots between the extracted and reference constraints. 


To systematically evaluate ML/DL approaches, the text dataset is typically split up into training and testing using different ratios (e.g., 2/3, 8/10)~\cite{goossens2023extracting, halioui2018bioinformatic}, or an external (i.e., different from the training) dataset is used~\cite{lopez_declarative_process_discovery}. 
For nondeterministic ML/DL methods, average performances over a number of runs (typically 5 or 10) are presented~\cite{goossens2023extracting}.
Halioui et al.~\cite{halioui2018bioinformatic} perform a 10-fold cross validation to establish optimal parameters;
Bellan et al.~\cite{bellan_pet_23} and Neuberger et al.~\cite{neuberger2023beyond} perform 5-fold cross-validation and take the average performance.\\
[0.2cm]
\textbf{Expert-based Evaluations}\\
Sonbol et al.~\cite{sonbol2023machine} asked experts to assess the understandability and usefulness of the extracted process model. 
Azevedo et al.~\cite{azevedo2018bpmn} asked 48 participants to assess whether the knowledge represented by the extracted process model is equivalent to the textual description.
Similarly, Ferreira et al.~\cite{ferreira2017semi} asked experts whether they agreed with  the extracted process models, using a 5-item Likert scale (from ``Fully Agree'' to ``Disagree''). They proposed six process models to the experts; 2 modeled based on mapping rules, and 4 purposefully modeled incorrectly.
All of the above represent \textit{aggregate} evaluations, as they focus on the process model as a whole. 

In \textit{element-by-element} evaluations, authors ask experts their opinion on the individual elements. 
Goncalves et al.~\cite{goncalves2011let} evaluated the extracted processes using a human modeler who separated valid (useful) from noise (not useful) elements.
In addition, Ferreira et al.~\cite{ferreira2017semi} asked experts to manually identify process elements from text (3 sentences) and subsequently compare them with the extracted elements. 


\section{Discussion}
\label{sec:discussion}

We identified multiple challenges and trends related to automated process extraction. 
We further discuss preliminary work that has been done using LLMs in this field, as well as promising developments in LLM architectures.

\subsection{Complexity of Natural Language}
Clearly, the problem of complex natural language is not limited to process extraction,
but rather any domain that relies on NLP. 
To curb this complexity, the text input can be restricted. 
We discuss works that rely on restricted text input in Section~\ref{sec:nlp_text_input}. For example, regarding user stories, Nasiri et al.~\cite{nasiri2023automatic} mention that a user story often uses the following format type: ``As \textlangle Role\textrangle, I want to \textlangle Action\textrangle, so that \textlangle Benefit\textrangle''. To avoid ambiguity and lack of clarity of free text, Goncalves et al.~\cite{goncalves2011let} consider stories from input text that is structured as a set of scenarios.
Some papers mention simplifying assumptions in the text: e.g., Text2Dec~\cite{etikala2020text2dec} assumes the description is sequential, contains no irrelevant information or redundancies, and contains only 1 main decision.

Quishpi et al.~\cite{quishpi2021extracting} make the point that language patterns expressing \textit{decisions} is less variable than those for \textit{control flow} descriptions. Hence, decisional models may be an easier target for process extraction. 
Van der Aa~\cite{vda_extracting_declarative_process} pointed out additional challenges regarding \textit{declarative} process extraction. 
For instance, subtle differences in text (e.g., modality; ``can" instead of ``must") will lead to semantically different constraints (e.g., precedence vs. response); in imperative models, these would typically lead to the same sequential relation.
Text may also indicate the negation of a constraint (e.g., ``invoice cannot be paid"), or logical relations (e.g., ``can be approved or rejected"), which must similarly be reflected in declarative constraints. 


\subsection{Use of Deep Learning for NLP}
\label{sec:discussion_dl}
We have observed a paradigm shift over the last 5 years that coincides with the emergence of DL in NLP; authors are increasingly using DL models such as Transformers (e.g., BERT \cite{goossens2023extracting}) and LSTM \cite{han2020bps} in process extraction. Until now, 4 works have applied Large Language Models (LLM), as discussed in Section~\ref{sec:discussion_llm}.

Quishpi et al.~\cite{quishpi2021extracting} note that classical, rule-based AI approaches can still have advantages over DL systems: (1) training DL systems from scratch requires huge amounts of annotated training data, and the cost of producing such datasets may be higher than the cost of encoding expert knowledge into rules; and (2) DL systems, in contrast to rule-based systems, operate as black boxes, and it is difficult to tailor their behaviour to improve output in case of wrong answers. 

Indeed, regarding training data (1), Neuberger et al.~\cite{neuberger2023beyond} hypothesised that the reason the end-to-end DL model Jerex, which was trained from scratch, was outperformed, was that the training data (PET dataset) was not yet extensive enough. Regarding the tailoring of behavior (2), the same authors point out that rule-based systems are also hard to adapt to even minor changes in the data.

Goossens et al.~\cite{goossens2023extracting} found that a \textit{pre-trained} BERT model, fine-tuned on process extraction (i.e., not trained from scratch), outperforms ML methods and even a DL model trained from scratch. Hence, regarding training data (1), pre-trained models may obviate the need for large-scale training datasets. However, fine-tuning still requires training on a labeled dataset for adjusting parameters. 
Moreover, pre-trained models may have difficulty with highly domain-specific terminology such as found in scientific texts~\cite{halioui2018bioinformatic}.

Finally, authors have pointed out that certain DL models may be intrinsically more suited towards automated process extraction:

\begin{itemize}
    \item LSTMs are adept at handling sequences, making them more suitable for modeling processes. Goossens et al.~\cite{goossens2023extracting} evaluated a Bi-LSTM-CRF model, pointing out that their particular structure affords them better insights into sentences. Similarly, several authors leveraged Bi-LSTM to obtain sentence encodings~\cite{qian2020approach,han2020bps}. 
    \item Han et al.~\cite{han2020bps} leverage Ordered Neurons LSTM (ON-LSTM), pointing out that ordered neurons add a structure-oriented inductive bias that better supports hierarchical process models.
    \item Neuberger et al.~\cite{neuberger2023beyond} pointed out that Jerex~\cite{Eberts2021}, an end-to-end DL method, can identify the textual location of an entity; this allows obtaining its surrounding text for enriching BPMN activity labels.
\end{itemize}

\subsection{Lack of Consistently Used, Gold-standard, Accessible and Large-Scale Evaluation Datasets}
\label{sec:discussion_data}
We previously pointed out a number of observations regarding evaluation datasets (Section~\ref{sec:eval_datasets}, Main Observations). We discuss these further below:

\vspace{0.2cm}
\noindent (1) \textit{There is a lack of consistently used evaluation datasets}.
Almost every study reviewed in this paper utilised a different dataset in their evaluation. Only the Friedrich~\cite{friedrich2010automated} and PET~\cite{bellan2022process} datasets are re-used, but only in 2 other studies each time; also, per dataset, only 1 of these 2 studies was not authored by its creators.
This greatly complicates an objective, third-party comparative evaluation. 
There are multiple potential reasons for this: the fact that many evaluation datasets can no longer be found online (issue (2)), or that there is lack of gold standard datasets (issue (3)).

\vspace{0.2cm}
\noindent (2) \textit{Many online datasets are no longer available.} While multiple articles provide links to their evaluation dataset, 4/9 of these links no longer work\footnote{Of course, this is a known problem beyond the field of automated process extraction.}, which prevents their re-use by third parties. 
We thus recommend making datasets available using immutable records, e.g., using Zenodo\footnote{https://zenodo.org/}, as opposed to university resources or even code repositories such as github\footnote{Unfortunately, the use of such repositories is not a solution, as 3/6 of those links no longer work either.}. This could be made a requirement by journals and conferences; multiple journals already have a data availability requirement.

\vspace{0.2cm}
\noindent (3) \textit{There is a paucity of gold-standard datasets}. There is currently 1 gold-standard dataset for control flow process extraction, namely the PET dataset (Section~\ref{sec:eval_datasets}). 
However, this dataset has only been available since 2022; 
moreover, it seems geared towards \textit{imperative} process models, as the manual annotations follow imperative model notations. 
For instance, annotated process elements include (AND/OR) gateways and associated conditions, branches, and flow (e.g., sequential) relations.
It is thus an open question whether the PET dataset is amenable to declarative process extraction.
Moreover, to the best of our knowledge, there is no gold standard dataset for extracting decisional models. 

\vspace{0.2cm}
\noindent (4) \textit{Most evaluation datasets are small in size}, i.e., in the order of tens or hundreds of individual process descriptions.
There is a need for large-scale datasets, not just for robust evaluations, but also for training/fine-tuning and testing DL models~\cite{neuberger2023beyond}.

\vspace{0.2cm}
To resolve issues (3) and (4), Neuberger et al.~\cite{neuberger_data_augment} recently proposed the application of data augmentation techniques for automatically synthesizing natural text for process extraction.
The authors found that simple data augmentation techniques already improved the accuracy of ML-based NER and RE models trained on the augmented data.
Example techniques include the increasing of linguistic variability, introducing variations in span length, and changing the directionality of actor-performer-recipient relations.
Nevertheless, data augmentation techniques still rely on an initial domain dataset to synthesize additional samples.
To better valuate manual data curation efforts, some conferences 
currently organize tracks that accept resources (datasets, benchmarks, software, ...) and their descriptions as publications.
Finally, we note that LLMs (see below) may end up removing the need of large training/fine-tuning datasets.
Even then, however, there may be a need to compare their performance with prior ML/DL methods.

\subsection{Promise of Large Language Models}
\label{sec:discussion_llm}
Large Language Models (LLM) 
can perform a variety of natural language tasks 
without requiring a specialized training or fine-tuning step. 
Bellan et al.~\cite{bellan_gpt3_2022} note that LLMs ``push pre-training to the extreme'': after being pre-trained on huge amounts of online textual data, they are able to carry out a variety of tasks by performing only small-scale ``in-context learning''. The latter involves a \textbf{prompt} with task instructions, contextual knowledge, possibly accompanied by few-shot examples (i.e., input and corresponding output), and the natural language text. 
Hence, LLMs have a huge potential for automated process extraction, as scalable  datasets for training and fine-tuning are currently lacking (Section~\ref{sec:discussion_data}).

At the same time, the fact that LLM re trained purely on Internet data may lead to inadvertently learning and reproducing biases therein~\cite{gupta2024navigating}. 
In process extraction, for instance, this may lead to processes that perpetuate discrimination on gender, race, religion or age~\cite{straw2020artificial}. We note that this problem expands beyond LLM and apply to any pre-trained language models, such as BERT~\cite{bhardwaj2021investigating}; techniques have been studied to identify and mitigate such biases~\cite{straw2020artificial}.
 
 Below, we shortly summarize initial work on the use of LLM for process extraction (Section~\ref{sec:llm_current}). Afterwards, we discuss promising developments such as Retrieval-Augmented Generation (RAG) and Multi-Agent Systems (Section~\ref{sec:llm_novel}).

\subsubsection{Current LLM-based Process Extraction}
\label{sec:llm_current}
At the time of writing (August 2024), to the best of our knowledge, 4 peer-reviewed papers~\cite{bellan_gpt3_2022,grohs2023large,forell_modeling_meets_llm,kourani_process_modelling_with_llm} present experimental results LLM-enabled process extraction. This relative paucity of work is not unexpected, considering their novelty; the ``game-changing'' ChatGPT was only released at the end of 2022. We discuss these works below along a salient set of dimensions.
\vspace{0.25cm}

\textbf{Prompts.}
All reviewed approaches construct a prompt with the textual process description, task instructions, typically a few few-shot (positive and negative) examples, and context knowledge. 
Given the prompt, the LLM then generates a process model based on the natural text.
Task instructions can range from relatively simple instructions~\cite{bellan_gpt3_2022,grohs2023large,kourani_process_modelling_with_llm} (e.g., list all activities in the text~\cite{bellan_gpt3_2022}) to detailed procedures (e.g., how to identify places and transitions for Petri nets~\cite{forell_modeling_meets_llm}).
Context knowledge tends to depend on what output is targeted;
since Bellan et al.~\cite{bellan_gpt3_2022} aim to extract participants, activities, and relations, they define these elements in the prompt, as well as the process domain.
Grohs et al.~\cite{grohs2023large} target both declarative and imperative control flow models; to cope with the LLM output limit, they define concise output formats that can later be mapped to e.g., Declare and BPMN, respectively.
Forell et al~\cite{forell_modeling_meets_llm} provide a comprehensive definition of a Petri net 
and an output format (JSON). 
Kourani et al.~\cite{kourani_process_modelling_with_llm} target Partially Ordered Workflow Language (POWL) as output; 
they define POWL and its semantics, and a set of Python functions to generate the POWL (see below).

\textbf{LLM Methodology.}
Most authors separate the task of generating a final output format. As mentioned, Grohs et al.~\cite{grohs2023large} output a custom format that is later mapped to Declare or BPMN. 
Forell et al.~\cite{forell_modeling_meets_llm} include a second phase, where the LLM is asked to output a concrete format for adaptability to different modeling tools. 
Kourani et al.~\cite{kourani_process_modelling_with_llm} ask the LLM to output Python code, which is then executed for the safe generation of POWL models.

Bellan et al.~\cite{bellan_gpt3_2022} apply an incremental approach; an initial prompt asks the LLM to identify activities from the text; subsequent prompts include these activities and ask for participants in, and directly-follows relations between, the activities.
Kourani et al.~\cite{kourani_process_modelling_with_llm} have users engage with the LLM in an iterative error handling loop; refined LLM prompts detail the error and ask the LLM to fix it.

\textbf{Suitability of LLM for process extraction}.
Bellan et al.~\cite{bellan_gpt3_2022} found that it is feasible to extract activities and their participants, but encountered challenges with directly-follows relations (avg. precision of 0.28). 
Grohs et al.~\cite{grohs2023large} compare the performance of their approach with the pre-LLM work by Van der Aa et al.~\cite{vda_extracting_declarative_process}; they found that GPT4 yields equal or higher F1 scores. 
Kourani et al.~\cite{kourani_process_modelling_with_llm} found that, regarding choice of LLM, GPT-4 performed better than Gemini. While the evaluation showed the effectiveness of LLM, the authors also note that manual effort is needed during the error handling loop~\cite{kourani_process_modelling_with_llm}. 
Forell et al.~\cite{forell_modeling_meets_llm} performed a small scale evaluation that illustrated the ability of LLM to generate correct Petri nets; however, issues were reported with AND/OR split and joins. They also found that the non-deterministic nature of LLM caused inconsistent output across multiple runs, 
which can be an issue for robust process modeling.


\vspace{0.25cm}
A number of vision papers have illustrated the general promise of LLM in BPM~\cite{busch2023just,vidgof2023large,kampik2023large}. For instance, Busch et al.~\cite{busch2023just} 
pointed out studies demonstrating that the performance of in-context learning can be comparable to, or even better than, fine-tuned models. The authors discuss a research agenda that includes the appropriate representation of processes in prompts; e.g., this is studied in detail by Berti et al.~\cite{berti2023abstractions} for querying processes using natural language and LLM.

\subsubsection{Promising Developments in the Field of LLMs}
\label{sec:llm_novel}

This section describes two recent innovations in the LLM field, namely Retrieval-Augmented Generation (RAG) and Multi-Agent systems (LLM-MA), and their potential to improve process extraction performance.

\vspace{0.25cm}
\textbf{Retrieval-Augmented Generation (RAG)} \\
In a RAG setup, given the LLM prompt, an extra step first queries an external source (typically, a vector database) for data relevant to the prompt. This data is then provided, together with the prompt, to the LLM~\cite{zhao2024_rag_applic}. 
By additionally providing the LLM with external, contextually relevant data, 
RAG aims to improve the accuracy and up-to-dateness of LLM responses.

Relevant to process extraction, RAG has been studied in event extraction~\cite{du2022_r-gqa} and code generation from natural text~\cite{poesia2022_synchromesh,li2023_skcoder,li2023_acecoder}. 
Typically, RAG is applied to find external event/code samples with labels relevant to the given text; subsequently, these samples are provided as positive examples to the generative model. 

\textbf{Event extraction.}
Event extraction identifies sequences of events in text, including their ``arguments'' (related individuals, organizations, and locations)~\cite{du2022_r-gqa}. 
The authors note that the structure of event descriptions is often quite similar in terms of syntax and semantics. Hence, a RAG setup is proposed that retrieves relevant event samples, i.e., with labels similar to the given event description, and provide them to the generative model. 
Experimental results show that the proposed approach outperforms baseline approaches~\cite{du2022_r-gqa}.

\textbf{Code generation}. Process extraction can be considered as a code generation problem, i.e., where ``process code'' is generated from input text\footnote{E.g., types of process code includes JSON~\cite{forell_modeling_meets_llm}, custom formats~\cite{grohs2023large}, and Python~\cite{kourani_process_modelling_with_llm}.}.
Similar to event extraction, multiple works on code generation~\cite{poesia2022_synchromesh,li2023_skcoder,li2023_acecoder} use RAG to retrieve code samples with labels similar to the user prompt, 
and then provide them as positive examples. 
This follows human developers' tendency to base themselves on code snippets that are relevant, but perhaps not identical, to their problem~\cite{li2023_skcoder,li2023_acecoder}. 
In this vein, SkCoder~\cite{li2023_skcoder} retrieves code snippets related to a textual description, and subsequently extracts a ``sketch'' (i.e., code skeleton) that retains only the code relevant to the given description.
To cope with the maximum input length of LLM, AceCoder~\cite{li2023_acecoder} applies a selector that filters out redundant code samples.
Synchromesh~\cite{poesia2022_synchromesh} applies Target Similarity Tuning to select semantically relevant samples
; Constrained Semantic Decoding is then used to generate only valid programs. Experiments show that the RAG setup can improve LLM accuracy and code validity~\cite{poesia2022_synchromesh}, and can outperform baseline LLM and non-RAG approaches~\cite{li2023_skcoder,li2023_acecoder}.

We propose that similar setups can be used to improve process extraction. 
Such a setup could leverage a repository of process descriptions, i.e., structured processes labeled with a textual description, to provide relevant samples to the generative model.
Moreover, an extra vector database can include word embeddings on organizational terminology; 
e.g., generated by a pre-trained BERT~\cite{devlin_bert_2018} model fine-tuned on such terminology for activities, roles and objects. 
Given an input text, an intermediate step can first extract individual terms and search the vector database for similar organizational terms;  
in the retrieval step, this terminology is then used to retrieve relevant samples from the process repository.

\vspace{0.25cm}
\textbf{Multi-Agent Systems (LLM-MA)}\\
In the code generation field, LLM-based Multi-Agent (LLM-MA) frameworks involve multiple LLM agents, each assigned a different role (e.g., analyst, coder, reviewer, tester). 
These agents then collaborate throughout the software generation lifecycle to generate the code output~\cite{guo_large_2024}. This setup draws inspiration from human software development, where complexity is managed by delegating tasks to team members, who then collaborate on producing the software. 
In the same vein, LLM-MA aim to improve the ability of LLM to manage the complexity of coding tasks.

Dong et al.~\cite{dong_self-collaboration_2024} present an LLM-MA framework where LLM agents respectively operate as analyst, coder, and tester; agents are instructed on how to collaborate and interact to generate code. In a sequence of stages, agents pass on their outputs to other agents (e.g., analyst to coder; tester back to coder). 
Qian et al.~\cite{qian_chatdev_2024} introduce ChatDev, an LLM-MA framework where agents assume e.g., programmer, reviewer, and tester roles. The code generation process is broken into a sequence of phases and subtasks. To solve a subtask, in a multi-turn dialogue, an instructor agent (e.g., reviewer) instructs an assistant agent (e.g., programmer), who responds with a potential solution  (e.g., fixing an endless loop). The solution from a subtask is passed on to the next subtask. In a “de-hallucination” mechanism, an assistant can seek clarification from the instructor before providing a solution. In both works, the authors found that their LLM-MA setup improved code generation performance.

We propose that a similar LLM-MA setup can be applied to process extraction. Such an LLM-MA would feature specialized agents, each dedicated to specific aspects of process extraction (e.g., identifying actors, activities, and objects, and relations between them). In a continuous feedback loop, a “critic” agent can iteratively evaluate their output, and, if needed, ask for refinements. An “integration” agent could then synthesize the outputs into a format such as e.g., BPMN, Petri nets, or Declare.

\section{Threats to Validity and Limitations}
\label{sec:threats_limits}

We acknowledge the existence of the following limitations and threats to validity:
\begin{itemize}
    \item \textit{Search scope}: Our literature search's effectiveness is invariably based on the chosen keywords (Table~\ref{table:synonyms_search_queries}), which were selected to target both NLP methods and process extraction. 
    \item \textit{Non-peer-reviewed and non-English articles}: Our reliance on peer-reviewed sources written in English may exclude insights from grey literature, such as technical reports, theses or corporate white papers, and non-English sources.
    For instance, this may overlook advances made outside of academic circles. 
    \item \textit{Coding of papers}: both authors (WV, SM) read and coded (i.e., categorised) the papers. Any inconsistencies were resolved through internal discussions.
    \item \textit{Rapidly evolving field}: the field of NLP is progressing rapidly, 
    especially regarding LLM-related work. 
    Recent developments may be reported in non-peer-reviewed channels (e.g., preprints from repositories like arXiv). 
    Of course, publications after the preparation of this manuscript (August 2024) will also not be included.
\end{itemize}

\section{Conclusion}
\label{sec:conclusion}

We presented a systematic literature review, conducted using the PRISMA guidelines, on the field of NLP-enabled process extraction. 
We cover publications up to 2023 that include rule-based, ML and DL methods, and with targets including control flow (imperative and declarative) and decisional models.
Review themes include natural language analysis, process model generation, and evaluation. 

Below, we summarise our answers to our research questions:\\
[0.2cm]
\indent \textbf{RQ1}: Which NLP and process generation methods, including publicly available tools, have been studied and used for extracting process models?\\
[0.1cm]
We presented a detailed landscape of process extraction methods,
their NLP and process model generation components, and public tools (Sections~\ref{sec:results_nlp}-~\ref{sec:results_eval}). 
As shown in Figure~\ref{fig:nlp_method_year}, historically (2011-2022), most approaches have used rule-based methods for NLP (14) compared to ML (6); DL methods started appearing after the advent of BERT in 2018 (5).
At the time of submission, we found initial experimental results (4) on the use of LLM (Section~\ref{sec:llm_current}).
Finally, we point out an imbalance in the type of target process model, 
with most works targeting imperative control flow models (15), followed at a distance by decisional (4) and declarative control flow models (3).
\\
[0.2cm]
\indent \textbf{RQ2}: To what extent have ML/DL methods been studied in process extraction?\\
[0.1cm]
Figure~\ref{fig:nlp_method_year} shows that, at this point (2023), the interest in ML/DL is on par with rule-based methods.
Table~\ref{tab:ml-dl} summarises the ML/DL models used for NLP tasks. 
We further discussed experimental results that show performance benefits of ML/DL over rule-based methods for NLP (Section~\ref{sec:nlp_computing}). 
We found that DL-based Transformers (BERT), which revolutionised NLP in 2018, were successfully applied to improve process extraction. 
We outlined ``non-traditional'' DL pipelines for NLP, and the DL models involved (Section~\ref{sec:nlp_computing}).
In our discussion, we summarize the observed use of DL for process extraction (Section~\ref{sec:discussion_dl}); including the suitability of certain DL models for downstream process extraction (e.g., LSTM) highlighted by study authors.\\
[0.2cm]
\indent \textbf{RQ3}: Which evaluation methods and datasets have been utilised to evaluate process extraction?\\
[0.1cm]
We provide a detailed listing of datasets used in the reviewed papers (Section~\ref{sec:eval_datasets}).
We hereby observed a barrier towards objective evaluation, namely the lack of accessible, scalable and gold-standard datasets for different types of process models.
We further presented a breakdown of orthogonal sets of evaluation methods, and classified the reviewed papers accordingly (Section~\ref{sec:eval_methods}).\\
[0.2cm]
Finally, we discussed the more recent LLM revolution as an opportunity to further improve  process extraction (Section~\ref{sec:discussion_llm}). We shortly reviewed initial work here, which is promising but also points out current challenges; and discussed promising developments (RAG, LLM-MA) that may be amenable to process extraction.
This systematic review of the ``pre-LLM'' era thus comes at an opportune moment:
for practitioners who want to apply process extraction, we present a detailed overview of state-of-the-art methods that can be used, including public NLP tools.
For researchers aiming to contribute to process extraction, we present the following:
\begin{itemize}
\item A review that describes a categorization of, and comparisons between, rule-based and ML/DL methods, non-traditional DL pipelines, and the use of LLM for process extraction. 
New studies can be inspired by this work, re-use their models, and/or compare their experimental results.
\item A comprehensive analysis of evaluation methods and datasets for process extraction, which can be directly re-used in new research.
\end{itemize}



\section{Acknowledgements} 
We would like to thank Daniel Amyot for a thorough proofreading of the paper, and Alireza Houshidari for providing insights on novel LLM setups.

\section{Declaration of Interest}
The authors report there are no competing interests to declare.

\section{Funding}
This work was supported by the Telfer School of Management, University of Ottawa, under a School of Management Research Grant (SMRG).

\newpage

\appendix
\setcounter{table}{0}
\renewcommand{\thetable}{A\arabic{table}}

\section{Detailed categorization of papers along classification themes.}

Regarding the NLP theme, we note that the vast majority of papers did not mention a restriction on text input (Section~\ref{sec:nlp_text_input}). We refer to Table~\ref{tab:nlp_tasks} for typical NLP pipeline tasks; Table~\ref{tab:ml-dl} for the use of ML/DL for these NLP tasks (NLP Computing Paradigm theme); and Table~\ref{table:nlp_tools} for public NLP tools utilized in the papers (NLP Tools theme). 

Below, we categorize the papers along the remaining Metadata (Table~\ref{tab:cat_metadata}), Process Model Generation (Table~\ref{tab:cat_proc_model_gen}), and Evaluation (Table~\ref{tab:cat_eval}) themes.

    \begin{longtable}{|p{8.5cm}|l|l|}
    \caption{Categorization of papers along Metadata themes.}
    \label{tab:cat_metadata}
    \\ \hline
        \textbf{Title} & \textbf{Authors} & \textbf{Year} \\ \hline
        A Machine Translation Like Approach to Generate Business Process Model from Textual Description & Sonbol et al. \cite{sonbol2023machine} & 2023 \\ \hline
        Extracting decision model components from natural language text for automated business decision modelling & Etikala~\cite{etikala2021extracting} & 2021 \\ \hline
        Extracting Decision Model and Notation models from text using deep learning techniques & Goossens et al. \cite{goossens2023extracting} & 2023 \\ \hline
        Text2Dec: Extracting Decision Dependencies from Natural Language Text for Automated DMN Decision Modelling & Etikala et al. \cite{etikala2020text2dec} & 2020 \\ \hline
        Generating BPMN diagram from textual requirements & Sholiq et al. \cite{sholiq2022generating} & 2022 \\ \hline
        Automatic generation of business process models from user stories & Nasiri et al. \cite{nasiri2023automatic} & 2023 \\ \hline
        BPMN model and text instructions automatic synchronization & Azevedo et al. \cite{azevedo2018bpmn} & 2018 \\ \hline
        A semi-automatic approach to identify business process elements in natural language texts & Ferreira et al. \cite{ferreira2017semi} & 2017 \\ \hline
        Let me tell you a story - on how to build process models & Goncalves et al. \cite{goncalves2011let} & 2011 \\ \hline
        Process model generation from natural language text & Friedrich et al. \cite{friedrich2011process} & 2011 \\ \hline
        PEWP: Process extraction based on word position in documents & Chen et al. \cite{chen2014pewp} & 2014 \\ \hline
        An Approach for Process Model Extraction by Multi-grained Text Classification & Qian et al. \cite{qian2020approach} & 2020 \\ \hline
        Extracting Decision Models from Textual Descriptions of Processes & Quishpi et al. \cite{quishpi2021extracting} & 2021 \\ \hline
        Bioinformatic Workflow Extraction from Scientific Texts based on Word Sense Disambiguation & Halioui et al. \cite{halioui2018bioinformatic} & 2018 \\ \hline
        A-BPS: automatic business process discovery service using ordered neurons LSTM & Han et al. \cite{han2020bps} & 2020 \\ \hline
        Beyond Rule-Based Named Entity Recognition and Relation Extraction for Process Model Generation from Natural Language Text & Neuberger et al. \cite{neuberger2023beyond} & 2023 \\ \hline
        A concept for generating business process models from natural language description & Honkisz et al. \cite{honkisz2018concept} & 2018 \\ \hline
        Extracting Declarative Process Models from Natural Language & Van der Aa et al. \cite{vda_extracting_declarative_process} & 2019 \\ \hline
        Declarative Process Discovery: Linking Process and Textual Views & L\'opez et al. \cite{lopez_declarative_process_discovery} & 2021 \\ \hline
        Assisted declarative process creation from natural language descriptions & L\'opez et al. \cite{lopez_assisted_declarative_process} & 2019 \\ \hline
\end{longtable}

\newpage
\begin{longtable}{|p{2cm}V{2}p{4cm}|p{3.2cm}|p{3.8cm}|}
    \caption{Categorization of papers along Process Model Generation themes.}
    \label{tab:cat_proc_model_gen}
    \\ \hline
        \multirow{2}{*}{\textbf{Citation}} & \multicolumn{3}{c|}{\textbf{Process Model Generation}} \\ \cline{2-4}
        ~ & \textit{Computing paradigm*} & \textit{Interm. repr.} & \textit{Target model} \\ \hline
        Sonbol et al. \cite{sonbol2023machine} & Knowledge-based - \newline Custom algorithms & Graph-based & Control flow - \newline Imperative (BPMN) \\ \hline
        Etikala\cite{etikala2021extracting} & Knowledge-based - \newline Templates / patterns & Dependency tuples & Decisional (DMN) \\ \hline
        Goossens et al. \cite{goossens2023extracting} & ML / DL - \newline (see Table~\ref{tab:ml-dl})\footnote{Process extraction is cast as an NLP NER problem.} & Dependency tuples & Decisional (DMN) 
\\ \hline
        Etikala et al. \cite{etikala2020text2dec} & Knowledge-based - \newline Templates / patterns & Dependency tuples & Decisional (DMN) \\ \hline
        Sholiq et al. \cite{sholiq2022generating} & Knowledge-based - \newline High-level rules & / & Control flow - \newline Imperative (BPMN) \\ \hline
        Nasiri et al. \cite{nasiri2023automatic} & Knowledge-based - \newline Concrete rules & / & Control flow - \newline Imperative (UML) \\ \hline
        Azevedo et al. \cite{azevedo2018bpmn} & Knowledge-based - \newline Custom & Tree-based & Control flow - \newline Imperative (BPMN) \\ \hline
        Ferreira et al. \cite{ferreira2017semi} & Knowledge-based - \newline Templates / patterns & / & Control flow - \newline Imperative (BPMN) \\ \hline
        Goncalves et al. \cite{goncalves2011let} & Knowledge-based - \newline High-level rules & / & Control flow - \newline Imperative (BPMN) \\ \hline
        Friedrich et al. \cite{friedrich2011process} & Knowledge-based - \newline Custom & Graph-based & Control flow - \newline Imperative (BPMN) \\ \hline
        Chen et al. \cite{chen2014pewp} & Knowledge-based - \newline Custom algorithms & / & Control flow - \newline Imperative (Custom) \\ \hline
        Qian et al. \cite{qian2020approach} & ML / DL - \newline (see Table~\ref{tab:ml-dl})\footnote{Process extraction is cast as an NLP text classification problem.}
& / & Control flow - \newline Imperative (BPMN) \\ \hline
        Quishpi et al. \cite{quishpi2021extracting} & Knowledge-based - \newline Templates / patterns & / & Decisional (DMN) \\ \hline
        Halioui et al. \cite{halioui2018bioinformatic} & Knowledge-based - \newline Concrete rules & / & Control flow - \newline 
Imperative (Custom) \\ \hline
        Han et al. \cite{han2020bps} & ML / DL - \newline (see Table~\ref{tab:ml-dl})\footnote{A process structure is extracted from the trained ON-LSTM model.}
& / & Control flow - \newline 
Imperative (BPMN) \\ \hline
        Neuberger et al. \cite{neuberger2023beyond} & ML / DL - \newline (see Table~\ref{tab:ml-dl})\footnote{Paper implements NER, RE, and coref. resol. for process extraction (latter is not elaborated).}
& / & Control flow - \newline 
Imperative (Unspec.) \\ \hline
        Honkisz et al. \cite{honkisz2018concept} & Knowledge-based - \newline Custom & Table & Control flow - \newline Imperative (BPMN) \\ \hline
        Van der Aa et al. \cite{vda_extracting_declarative_process} & Knowledge-based - \newline Custom algorithms & / & Control flow - \newline 
Declarative (Declare) \\ \hline
        L\'opez et al. \cite{lopez_declarative_process_discovery} & Knowledge-based - \newline Custom algorithms
& / & Control flow - \newline 
Declarative (DCR) \\ \hline
        L\'opez et al. \cite{lopez_assisted_declarative_process} & Knowledge-based - \newline Custom & / & Control flow - \newline 
Declarative (DCR) \\ \hline
\end{longtable}

\newpage
    \begin{longtable}{|p{2.5cm}V{2}p{4.1cm}|p{2.8cm}|p{3cm}|}
    \caption{Categorization of papers along Evaluation themes.}
    \label{tab:cat_eval}
    \\ \hline
    \multirow{2}{*}{\textbf{Citation}} & \multicolumn{3}{c|}{\textbf{Evaluation}*} \\ \cline{2-4}
        ~ & \textit{Type}** & \textit{Metrics}$\dagger$ & \textit{Dataset}$\ddagger$ \\ \hline
        Sonbol et al. \cite{sonbol2023machine} & Holistic - Systematic
(Graph-based); \newline 
Holistic - 
Expert-based & GED
& Friedrich \cite{friedrich2010automated} \\ \hline
        Etikala \cite{etikala2021extracting} & Component-based - \newline Systematic & P, R & N/A \\ \hline
        Goossens et al. \cite{goossens2023extracting} & Component-based - \newline Systematic; \newline 
Holistic - Systematic & P, R, F1 & N/A \\ \hline
        Etikala et al. \cite{etikala2020text2dec} & Holistic - Systematic & N/A & Listed in paper \\ \hline
        Sholiq et al. \cite{sholiq2022generating} & Holistic - Systematic & MRE
        & Listed in paper \\ \hline
        Nasiri et al. \cite{nasiri2023automatic} & Holistic - Systematic & P, R, Acc & Listed in paper \\ \hline
        Azevedo et al. \cite{azevedo2018bpmn} & Holistic - Expert-based & 
        0-100 \& Likert scale & N/A \\ \hline
        Ferreira et al. \cite{ferreira2017semi} & Holistic - Systematic; \newline 
Holistic - 
Expert-based & P, R, F1, Acc; \newline Likert scale & N/A \\ \hline
        Goncalves et al. \cite{goncalves2011let} & Holistic - 
Expert-based & \# invalid \& \newline total elements \newline extracted & N/A \\ \hline
        Friedrich et al. \cite{friedrich2011process} & Holistic - Systematic
(Graph-based) & GED
& Friedrich \cite{friedrich2010automated} \\ \hline
        Chen et al. \cite{chen2014pewp} & Holistic - Systematic & NDPM
        , R & N/A \\ \hline
        Qian et al. \cite{qian2020approach} & Component-based - \newline Systematic; \newline 
Holistic - Systematic (Aggregate) & Acc & N/A \\ \hline
        Quishpi et al. \cite{quishpi2021extracting} & Holistic - Systematic & P, R, F1 & N/A \\ \hline
        Halioui et al. \cite{halioui2018bioinformatic} & Holistic - Systematic & P, R, F1 & {\footnotesize \href{https://data.world/growler/datasets}{https://data.world/ growler/datasets}} \\ \hline
        Han et al. \cite{han2020bps} & Holistic - Systematic
(Graph-based) & Graph similarity score & N/A \\ \hline
        Neuberger et al. \cite{neuberger2023beyond} & Component-based - \newline Systematic; \newline 
Holistic - Systematic & P, R, F1 & {\footnotesize \href{http://huggingface.co/datasets/patriziobellan/PET}{http://huggingface.co/ datasets/patriziobellan/ PET}} \\ \hline
        Honkisz et al. \cite{honkisz2018concept} & Holistic - Systematic & N/A & Listed in paper \\ \hline
        Van der Aa et al. \cite{vda_extracting_declarative_process} & Holistic - Systematic & P, R, F1 & {\footnotesize \href{https://github.com/hanvanderaa/declareextraction}{https://github.com/ \newline hanvanderaa/\newline declareextraction}} \\ \hline
        L\'opez et al. \cite{lopez_declarative_process_discovery} & Holistic - Systematic & P, R, F1 & N/A \\ \hline
        L\'opez et al. \cite{lopez_assisted_declarative_process} & Holistic - Systematic & N/A & N/A \\ \hline
\multicolumn{4}{p{14cm}}{* 
P=Precision, R=Recall, Acc=Accuracy, F1=F1-score.} \\
\multicolumn{4}{p{14cm}}{** `Systematic' evaluations are `element-by-element' unless otherwise specified.} \\
\multicolumn{4}{p{14cm}}{$\dagger$ `N/A' when evaluation results are narratively described.} \\
\multicolumn{4}{p{14cm}}{$\ddagger$ `N/A' when the dataset is private \textit{or} the link was not working at submission.} \\
\end{longtable}


 \bibliographystyle{elsarticle-num} 
 \bibliography{cas-refs}

\begin{thebibliography}{10}
\expandafter\ifx\csname url\endcsname\relax
  \def\url#1{\texttt{#1}}\fi
\expandafter\ifx\csname urlprefix\endcsname\relax\def\urlprefix{URL }\fi
\expandafter\ifx\csname href\endcsname\relax
  \def\href#1#2{#2} \def\path#1{#1}\fi

\bibitem{weske2007business}
M.~Weske, Business process management--concepts, languages, architectures, verlag, Berlin (2007).

\bibitem{bellan2020qualitative}
P.~Bellan, M.~Dragoni, C.~Ghidini, A qualitative analysis of the state of the art in process extraction from text., DP@ AI* IA (2020) 19--30.

\bibitem{wvda_declare}
W.~M.~P. van~der Aalst, M.~Pesic, H.~Schonenberg, Declarative workflows: Balancing between flexibility and support, Computer Science - Research and Development 23 (2009) 99--113.
\newblock \href {https://doi.org/10.1007/s00450-009-0057-9} {\path{doi:10.1007/s00450-009-0057-9}}.

\bibitem{van2018challenges}
H.~Van~der Aa, J.~Carmona~Vargas, H.~Leopold, J.~Mendling, L.~Padr{\'o}, Challenges and opportunities of applying natural language processing in business process management, in: COLING 2018: The 27th International Conference on Computational Linguistics: Proceedings of the Conference: August 20-26, 2018 Santa Fe, New Mexico, USA, Association for Computational Linguistics, 2018, pp. 2791--2801.

\bibitem{Davenport2019}
T.~H. Davenport, J.~Guszcza, T.~Smith, B.~Stiller, \href{https://www2.deloitte.com/us/en/insights/topics/analytics/insight-driven-organization.html}{Insight-driven organization | deloitte insights} (2019).
\newline\urlprefix\url{https://www2.deloitte.com/us/en/insights/topics/analytics/insight-driven-organization.html}

\bibitem{maqbool2019comprehensive}
B.~Maqbool, F.~Azam, M.~W. Anwar, W.~H. Butt, J.~Zeb, I.~Zafar, A.~K. Nazir, Z.~Umair, A comprehensive investigation of bpmn models generation from textual requirements—techniques, tools and trends, in: Information Science and Applications 2018: ICISA 2018, Springer, 2019, pp. 543--557.

\bibitem{bordignon2018natural}
A.~C. d.~A. Bordignon, L.~H. Thom, T.~S. Silva, V.~S. Dani, M.~Fantinato, R.~C.~B. Ferreira, Natural language processing in business process identification and modeling: a systematic literature review, Simp{\'o}sio Brasileiro de Sistemas de Informa{\c{c}}{\~a}o (SBSI) (2018).

\bibitem{herbst_inductive_1999}
J.~Herbst, D.~Karagiannis, An inductive approach to the acquisition and adaptation of workflow models, in: Proceedings of the {IJCAI}, 1999, pp. 52--57.

\bibitem{goossens2023extracting}
A.~Goossens, J.~De~Smedt, J.~Vanthienen, Extracting decision model and notation models from text using deep learning techniques, Expert Systems with Applications 211 (2023) 118667.

\bibitem{han2020bps}
X.~Han, L.~Hu, L.~Mei, Y.~Dang, S.~Agarwal, X.~Zhou, P.~Hu, A-bps: automatic business process discovery service using ordered neurons lstm, in: 2020 IEEE International Conference on Web Services (ICWS), IEEE, 2020, pp. 428--432.

\bibitem{kampik2023large}
T.~Kampik, C.~Warmuth, A.~Rebmann, R.~Agam, L.~N. Egger, A.~Gerber, J.~Hoffart, J.~Kolk, P.~Herzig, G.~Decker, et~al., Large process models: Business process management in the age of generative ai, arXiv preprint arXiv:2309.00900 (2023).

\bibitem{bellan_gpt3_2022}
P.~Bellan, M.~Dragoni, C.~Ghidini, Extracting business process entities and relations from text using pre-trained language models and in-context learning, in: J.~P.~A. Almeida, D.~Karastoyanova, G.~Guizzardi, M.~Montali, F.~M. Maggi, C.~M. Fonseca (Eds.), Enterprise Design, Operations, and Computing, Springer International Publishing, Cham, 2022, pp. 182--199.

\bibitem{grohs2023large}
M.~Grohs, L.~Abb, N.~Elsayed, J.-R. Rehse, Large language models can accomplish business process management tasks, arXiv preprint arXiv:2307.09923 (2023).

\bibitem{forell_modeling_meets_llm}
M.~Forell, S.~Schüler, Modeling meets large language models, Modellierung 2024 Satellite Events (2024).
\newblock \href {https://doi.org/10.18420/modellierung2024-ws-003} {\path{doi:10.18420/modellierung2024-ws-003}}.

\bibitem{kourani_process_modelling_with_llm}
H.~Kourani, A.~Berti, D.~Schuster, W.~M.~P. van~der Aalst, Process modeling with large language models, in: H.~van~der Aa, D.~Bork, R.~Schmidt, A.~Sturm (Eds.), Enterprise, Business-Process and Information Systems Modeling, Springer Nature Switzerland, Cham, 2024, pp. 229--244.

\bibitem{bellan2021process}
P.~Bellan, M.~Dragoni, C.~Ghidini, Process extraction from text: state of the art and challenges for the future, arXiv preprint arXiv:2110.03754 (2021).

\bibitem{devlin_bert_2018}
J.~Devlin, M.-W. Chang, K.~Lee, K.~Toutanova, {BERT}: {Pre}-training of {Deep} {Bidirectional} {Transformers} for {Language} {Understanding}, http://arxiv.org/abs/1810.04805 (Oct. 2018).
\newblock \href {https://doi.org/10.48550/arxiv.1810.04805} {\path{doi:10.48550/arxiv.1810.04805}}.

\bibitem{riefer2016mining}
M.~Riefer, S.~F. Ternis, T.~Thaler, Mining process models from natural language text: A state-of-the-art analysis, Multikonferenz Wirtschaftsinformatik (MKWI-16), March (2016) 9--11.

\bibitem{omg_bpmn}
{Object Management Group (OMG)}, \href{https://www.bpmn.org/}{Business process model and notation (bpmn)} (2024).
\newline\urlprefix\url{https://www.bpmn.org/}

\bibitem{schuler_state_2024}
S.~Sch{\"u}ler, S.~Alpers, State of the art: Automatic generation of business process models, in: J.~De~Weerdt, L.~Pufahl (Eds.), Business Process Management Workshops, Springer Nature Switzerland, Cham, 2024, pp. 161--173.

\bibitem{Page2021}
M.~J. Page, J.~E. McKenzie, P.~M. Bossuyt, I.~Boutron, T.~C. Hoffmann, C.~D. Mulrow, L.~Shamseer, J.~M. Tetzlaff, E.~A. Akl, S.~E. Brennan, R.~Chou, J.~Glanville, J.~M. Grimshaw, A.~Hr{\'o}bjartsson, M.~M. Lalu, T.~Li, E.~W. Loder, E.~Mayo-Wilson, S.~McDonald, L.~A. McGuinness, L.~A. Stewart, J.~Thomas, A.~C. Tricco, V.~A. Welch, P.~Whiting, D.~Moher, \href{https://www.bmj.com/content/372/bmj.n71}{The prisma 2020 statement: an updated guideline for reporting systematic reviews}, BMJ 372 (2021).
\newblock \href {http://arxiv.org/abs/https://www.bmj.com/content/372/bmj.n71.full.pdf} {\path{arXiv:https://www.bmj.com/content/372/bmj.n71.full.pdf}}, \href {https://doi.org/10.1136/bmj.n71} {\path{doi:10.1136/bmj.n71}}.
\newline\urlprefix\url{https://www.bmj.com/content/372/bmj.n71}

\bibitem{Fereday2006}
J.~Fereday, E.~Muir-Cochrane, \href{http://journals.sagepub.com/doi/10.1177/160940690600500107}{Demonstrating {Rigor} {Using} {Thematic} {Analysis}: {A} {Hybrid} {Approach} of {Inductive} and {Deductive} {Coding} and {Theme} {Development}}, International Journal of Qualitative Methods 5~(1) (2006) 80--92, publisher: SAGE PublicationsSage CA: Los Angeles, CA.
\newblock \href {https://doi.org/10.1177/160940690600500107} {\path{doi:10.1177/160940690600500107}}.
\newline\urlprefix\url{http://journals.sagepub.com/doi/10.1177/160940690600500107}

\bibitem{nasiri2023automatic}
S.~Nasiri, A.~Adadi, M.~Lahmer, Automatic generation of business process models from user stories, International Journal of Electrical and Computer Engineering 13~(1) (2023) 809.

\bibitem{goncalves2011let}
J.~C. d.~A. Goncalves, F.~M. Santoro, F.~A. Bai{\~a}o, Let me tell you a story-on how to build process models, Journal of Universal Computer Science 17~(2) (2011) 276--295.

\bibitem{halioui2018bioinformatic}
A.~Halioui, P.~Valtchev, A.~B. Diallo, Bioinformatic workflow extraction from scientific texts based on word sense disambiguation, IEEE/ACM transactions on computational biology and bioinformatics 15~(6) (2018) 1979--1990.

\bibitem{de2008stanford}
M.-C. De~Marneffe, C.~D. Manning, Stanford typed dependencies manual, Tech. rep., Technical report, Stanford University (2008).

\bibitem{pennington2014glove}
J.~Pennington, R.~Socher, C.~D. Manning, Glove: Global vectors for word representation, in: Proceedings of the 2014 conference on empirical methods in natural language processing (EMNLP), 2014, pp. 1532--1543.

\bibitem{qian2020approach}
C.~Qian, L.~Wen, A.~Kumar, L.~Lin, L.~Lin, Z.~Zong, S.~Li, J.~Wang, An approach for process model extraction by multi-grained text classification, in: Advanced Information Systems Engineering: 32nd International Conference, CAiSE 2020, Grenoble, France, June 8--12, 2020, Proceedings 32, Springer, 2020, pp. 268--282.

\bibitem{neuberger2023beyond}
J.~Neuberger, L.~Ackermann, S.~Jablonski, Beyond rule-based named entity recognition and relation extraction for process model generation from natural language text, in: Cooperative Information Systems, Springer Nature Switzerland, Cham, 2023, pp. 179--197.

\bibitem{lopez_declarative_process_discovery}
H.~A. L{\'o}pez, R.~Str{\o}msted, J.-M. Niyodusenga, M.~Marquard, Declarative process discovery: Linking process and textual views, in: S.~Nurcan, A.~Korthaus (Eds.), Intelligent Information Systems, Springer International Publishing, Cham, 2021, pp. 109--117.

\bibitem{etikala2020text2dec}
V.~Etikala, Z.~Van~Veldhoven, J.~Vanthienen, Text2dec: extracting decision dependencies from natural language text for automated dmn decision modelling, in: International Conference on Business Process Management, Springer, 2020, pp. 367--379.

\bibitem{etikala2021extracting}
V.~Etikala, Extracting decision model components from natural language text for automated business decision modelling., in: RuleML+ RR (Supplement), 2021.

\bibitem{Eberts2021}
M.~Eberts, A.~Ulges, \href{https://api.semanticscholar.org/CorpusID:231879840}{An end-to-end model for entity-level relation extraction using multi-instance learning}, in: Conference of the European Chapter of the Association for Computational Linguistics, 2021, pp. 3650--3660.
\newline\urlprefix\url{https://api.semanticscholar.org/CorpusID:231879840}

\bibitem{bellan2022process}
P.~Bellan, C.~Ghidini, M.~Dragoni, S.~P. Ponzetto, H.~van~der Aa, Process extraction from natural language text: the pet dataset and annotation guidelines, in: Proceedings of the Sixth Workshop on Natural Language for Artificial Intelligence (NL4AI 2022) co-located with 21th International Conference of the Italian Association for Artificial Intelligence (AI* IA 2022), Vol. 3287, CEUR-WS. org, 2022, pp. 177--191.

\bibitem{sanh2020distilbert}
V.~Sanh, L.~Debut, J.~Chaumond, T.~Wolf, Distilbert, a distilled version of bert: smaller, faster, cheaper and lighter (2020).
\newblock \href {http://arxiv.org/abs/1910.01108} {\path{arXiv:1910.01108}}.

\bibitem{ezencan2020comparison}
A.~Ezen-Can, A comparison of lstm and bert for small corpus (2020).
\newblock \href {http://arxiv.org/abs/2009.05451} {\path{arXiv:2009.05451}}.

\bibitem{shen2018ordered}
Y.~Shen, S.~Tan, A.~Sordoni, A.~Courville, Ordered neurons: Integrating tree structures into recurrent neural networks, arXiv preprint arXiv:1810.09536 (2018).

\bibitem{li2005using}
Y.~Li, K.~Bontcheva, H.~Cunningham, Using uneven margins svm and perceptron for information extraction, in: Proceedings of the Ninth Conference on Computational Natural Language Learning (CoNLL-2005), 2005, pp. 72--79.

\bibitem{chen2014pewp}
Y.~Chen, Z.~Ding, H.~Sun, Pewp: Process extraction based on word position in documents, in: Ninth International Conference on Digital Information Management (ICDIM 2014), IEEE, 2014, pp. 135--140.

\bibitem{sintoris2017extracting}
K.~Sintoris, K.~Vergidis, Extracting business process models using natural language processing (nlp) techniques, in: 2017 IEEE 19th conference on business informatics (CBI), Vol.~1, IEEE, 2017, pp. 135--139.

\bibitem{ferreira2017semi}
R.~C.~B. Ferreira, L.~H. Thom, M.~Fantinato, A semi-automatic approach to identify business process elements in natural language texts, in: International Conference on Enterprise Information Systems, Vol.~2, SCITEPRESS, 2017, pp. 250--261.

\bibitem{honkisz2018concept}
K.~Honkisz, K.~Kluza, P.~Wi{\'s}niewski, A concept for generating business process models from natural language description, in: Knowledge Science, Engineering and Management: 11th International Conference, KSEM 2018, Changchun, China, August 17--19, 2018, Proceedings, Part I 11, Springer, 2018, pp. 91--103.

\bibitem{sonbol2023machine}
R.~Sonbol, G.~Rebdawi, N.~Ghneim, A machine translation like approach to generate business process model from textual description, SN Computer Science 4~(3) (2023) 291.

\bibitem{friedrich2011process}
F.~Friedrich, J.~Mendling, F.~Puhlmann, Process model generation from natural language text, in: Advanced Information Systems Engineering: 23rd International Conference, CAiSE 2011, London, UK, June 20-24, 2011. Proceedings 23, Springer, 2011, pp. 482--496.

\bibitem{vda_extracting_declarative_process}
H.~van~der Aa, C.~Di~Ciccio, H.~Leopold, H.~A. Reijers, Extracting declarative process models from natural language, in: P.~Giorgini, B.~Weber (Eds.), Advanced Information Systems Engineering, Springer International Publishing, Cham, 2019, pp. 365--382.

\bibitem{quishpi2021extracting}
L.~Quishpi, J.~Carmona, L.~Padr{\'o}, Extracting decision models from textual descriptions of processes, in: International Conference on Business Process Management, Springer, 2021, pp. 85--102.

\bibitem{sholiq2022generating}
S.~Sholiq, R.~Sarno, E.~S. Astuti, Generating bpmn diagram from textual requirements, Journal of King Saud University-Computer and Information Sciences 34~(10) (2022) 10079--10093.

\bibitem{lopez_assisted_declarative_process}
H.~A. L{\'o}pez, M.~Marquard, L.~Muttenthaler, R.~I. Str{\o}msted, \href{https://api.semanticscholar.org/CorpusID:208207164}{Assisted declarative process creation from natural language descriptions}, 2019 IEEE 23rd International Enterprise Distributed Object Computing Workshop (EDOCW) (2019) 96--99.
\newline\urlprefix\url{https://api.semanticscholar.org/CorpusID:208207164}

\bibitem{cunningham2013getting}
H.~Cunningham, V.~Tablan, A.~Roberts, K.~Bontcheva, Getting more out of biomedical documents with gate's full lifecycle open source text analytics, PLoS computational biology 9~(2) (2013) e1002854.

\bibitem{azevedo2018bpmn}
L.~G. Azevedo, R.~D.~A. Rodrigues, K.~Revoredo, Bpmn model and text instructions automatic synchronization., in: ICEIS (1), 2018, pp. 484--491.

\bibitem{omg_uml}
{Object Management Group (OMG)}, \href{https://www.omg.org/spec/UML/}{Unified modeling language (uml)} (2024).
\newline\urlprefix\url{https://www.omg.org/spec/UML/}

\bibitem{hildebrandt_dcr}
T.~Hildebrandt, R.~R. Mukkamala, Declarative event-based workflow as distributed dynamic condition response graphs, PLACES 69 (10 2011).
\newblock \href {https://doi.org/10.4204/EPTCS.69.5} {\path{doi:10.4204/EPTCS.69.5}}.

\bibitem{omg_dmn}
{Object Management Group (OMG)}, \href{https://www.omg.org/dmn/}{Decision {Model} and {Notation} ({DMN})} (2024).
\newline\urlprefix\url{https://www.omg.org/dmn/}

\bibitem{friedrich2010automated}
F.~Friedrich, Automated generation of business process models from natural language input, M. Sc., School of Business and Economics. Humboldt-Universit{\"a}t zu Berli (2010).

\bibitem{bellan_pet_23}
P.~Bellan, H.~van~der Aa, M.~Dragoni, C.~Ghidini, S.~P. Ponzetto, Pet: An annotated dataset for process extraction from natural language text tasks, in: C.~Cabanillas, N.~F. Garmann-Johnsen, A.~Koschmider (Eds.), Business Process Management Workshops, Springer International Publishing, Cham, 2023, pp. 315--321.

\bibitem{weidlich2011}
M.~Weidlich, J.~Mendling, M.~Weske, Efficient consistency measurement based on behavioral profiles of process models, IEEE Transactions on Software Engineering 37~(3) (2011) 410--429.
\newblock \href {https://doi.org/10.1109/TSE.2010.96} {\path{doi:10.1109/TSE.2010.96}}.

\bibitem{dijkman2011}
R.~Dijkman, M.~Dumas, B.~{van Dongen}, R.~Käärik, J.~Mendling, \href{https://www.sciencedirect.com/science/article/pii/S0306437910001006}{Similarity of business process models: Metrics and evaluation}, Information Systems 36~(2) (2011) 498--516, special Issue: Semantic Integration of Data, Multimedia, and Services.
\newblock \href {https://doi.org/https://doi.org/10.1016/j.is.2010.09.006} {\path{doi:https://doi.org/10.1016/j.is.2010.09.006}}.
\newline\urlprefix\url{https://www.sciencedirect.com/science/article/pii/S0306437910001006}

\bibitem{neuberger_data_augment}
J.~Neuberger, L.~Doll, B.~Engelmann, L.~Ackermann, S.~Jablonski, Leveraging data augmentation for process information extraction, in: H.~van~der Aa, D.~Bork, R.~Schmidt, A.~Sturm (Eds.), Enterprise, Business-Process and Information Systems Modeling, Springer Nature Switzerland, Cham, 2024, pp. 57--70.

\bibitem{gupta2024navigating}
A.~G. Brij B.~Gupta, V.~Arya, \href{https://doi.org/10.1080/17517575.2024.2310846}{Navigating the security landscape of large language models in enterprise information systems}, Enterprise Information Systems 18~(4) (2024) 2310846.
\newblock \href {http://arxiv.org/abs/https://doi.org/10.1080/17517575.2024.2310846} {\path{arXiv:https://doi.org/10.1080/17517575.2024.2310846}}, \href {https://doi.org/10.1080/17517575.2024.2310846} {\path{doi:10.1080/17517575.2024.2310846}}.
\newline\urlprefix\url{https://doi.org/10.1080/17517575.2024.2310846}

\bibitem{straw2020artificial}
I.~Straw, C.~Callison-Burch, \href{https://doi.org/10.1371/journal.pone.0240376}{Artificial intelligence in mental health and the biases of language based models}, PLOS ONE 15~(12) (2020) 1--19.
\newblock \href {https://doi.org/10.1371/journal.pone.0240376} {\path{doi:10.1371/journal.pone.0240376}}.
\newline\urlprefix\url{https://doi.org/10.1371/journal.pone.0240376}

\bibitem{bhardwaj2021investigating}
R.~Bhardwaj, N.~Majumder, S.~Poria, Investigating gender bias in bert, Cognitive Computation 13~(4) (2021) 1008--1018.

\bibitem{busch2023just}
K.~Busch, A.~Rochlitzer, D.~Sola, H.~Leopold, Just tell me: Prompt engineering in business process management, in: International Conference on Business Process Modeling, Development and Support, Springer, 2023, pp. 3--11.

\bibitem{vidgof2023large}
M.~Vidgof, S.~Bachhofner, J.~Mendling, Large language models for business process management: Opportunities and challenges, arXiv preprint arXiv:2304.04309 (2023).

\bibitem{berti2023abstractions}
A.~Berti, D.~Schuster, W.~M. van~der Aalst, Abstractions, scenarios, and prompt definitions for process mining with llms: A case study, arXiv preprint arXiv:2307.02194 (2023).

\bibitem{zhao2024_rag_applic}
P.~Zhao, H.~Zhang, Q.~Yu, Z.~Wang, Y.~Geng, F.~Fu, L.~Yang, W.~Zhang, J.~Jiang, B.~Cui, \href{https://arxiv.org/abs/2402.19473}{Retrieval-augmented generation for ai-generated content: A survey} (2024).
\newblock \href {http://arxiv.org/abs/2402.19473} {\path{arXiv:2402.19473}}.
\newline\urlprefix\url{https://arxiv.org/abs/2402.19473}

\bibitem{du2022_r-gqa}
X.~Du, H.~Ji, \href{https://aclanthology.org/2022.emnlp-main.307}{Retrieval-augmented generative question answering for event argument extraction}, in: Y.~Goldberg, Z.~Kozareva, Y.~Zhang (Eds.), Proceedings of the 2022 Conference on Empirical Methods in Natural Language Processing, Association for Computational Linguistics, Abu Dhabi, United Arab Emirates, 2022, pp. 4649--4666.
\newblock \href {https://doi.org/10.18653/v1/2022.emnlp-main.307} {\path{doi:10.18653/v1/2022.emnlp-main.307}}.
\newline\urlprefix\url{https://aclanthology.org/2022.emnlp-main.307}

\bibitem{poesia2022_synchromesh}
G.~Poesia, O.~Polozov, V.~Le, A.~Tiwari, G.~Soares, C.~Meek, S.~Gulwani, \href{https://arxiv.org/abs/2201.11227}{Synchromesh: Reliable code generation from pre-trained language models} (2022).
\newblock \href {http://arxiv.org/abs/2201.11227} {\path{arXiv:2201.11227}}.
\newline\urlprefix\url{https://arxiv.org/abs/2201.11227}

\bibitem{li2023_skcoder}
J.~Li, Y.~Li, G.~Li, Z.~Jin, Y.~Hao, X.~Hu, \href{https://doi.org/10.1109/ICSE48619.2023.00179}{Skcoder: A sketch-based approach for automatic code generation}, in: Proceedings of the 45th International Conference on Software Engineering, ICSE '23, IEEE Press, 2023, p. 2124–2135.
\newblock \href {https://doi.org/10.1109/ICSE48619.2023.00179} {\path{doi:10.1109/ICSE48619.2023.00179}}.
\newline\urlprefix\url{https://doi.org/10.1109/ICSE48619.2023.00179}

\bibitem{li2023_acecoder}
J.~Li, Y.~Zhao, Y.~Li, G.~Li, Z.~Jin, \href{https://arxiv.org/abs/2303.17780}{Acecoder: Utilizing existing code to enhance code generation} (2023).
\newblock \href {http://arxiv.org/abs/2303.17780} {\path{arXiv:2303.17780}}.
\newline\urlprefix\url{https://arxiv.org/abs/2303.17780}

\bibitem{guo_large_2024}
T.~Guo, X.~Chen, Y.~Wang, R.~Chang, S.~Pei, N.~V. Chawla, O.~Wiest, X.~Zhang, \href{https://doi.org/10.48550/arXiv.2402.01680}{Large {Language} {Model} based {Multi}-{Agents}: {A} {Survey} of {Progress} and {Challenges}}, CoRR abs/2402.01680, arXiv: 2402.01680 (2024).
\newblock \href {https://doi.org/10.48550/ARXIV.2402.01680} {\path{doi:10.48550/ARXIV.2402.01680}}.
\newline\urlprefix\url{https://doi.org/10.48550/arXiv.2402.01680}

\bibitem{dong_self-collaboration_2024}
Y.~Dong, X.~Jiang, Z.~Jin, G.~Li, \href{https://doi.org/10.1145/3672459}{Self-collaboration {Code} {Generation} via {ChatGPT}}, ACM Trans. Softw. Eng. Methodol.Place: New York, NY, USA Publisher: Association for Computing Machinery (Jun. 2024).
\newblock \href {https://doi.org/10.1145/3672459} {\path{doi:10.1145/3672459}}.
\newline\urlprefix\url{https://doi.org/10.1145/3672459}

\bibitem{qian_chatdev_2024}
C.~Qian, W.~Liu, H.~Liu, N.~Chen, Y.~Dang, J.~Li, C.~Yang, W.~Chen, Y.~Su, X.~Cong, J.~Xu, D.~Li, Z.~Liu, M.~Sun, \href{https://aclanthology.org/2024.acl-long.810}{{ChatDev}: {Communicative} {Agents} for {Software} {Development}}, in: L.-W. Ku, A.~Martins, V.~Srikumar (Eds.), Proceedings of the 62nd {Annual} {Meeting} of the {Association} for {Computational} {Linguistics} ({Volume} 1: {Long} {Papers}), {ACL} 2024, {Bangkok}, {Thailand}, {August} 11-16, 2024, Association for Computational Linguistics, 2024, pp. 15174--15186.
\newline\urlprefix\url{https://aclanthology.org/2024.acl-long.810}

\end{thebibliography}






\end{document}